\documentclass[10pt,twocolumn,letterpaper]{article}

\usepackage{cvpr}
\usepackage{times}
\usepackage{amsmath}
\usepackage{amssymb}
\usepackage{booktabs}
\usepackage{graphicx}
\usepackage{subfig}
\usepackage{tabularx}
\usepackage{CJKutf8}

\newcommand{\eat}[1]{}

\makeatletter
\@namedef{ver@everyshi.sty}{}
\makeatother

\usepackage{tikz}
\usepackage{pgf, pgfsys, pgffor}
\usepackage{pgfplots}

\definecolor{rose}{rgb}{0.92, 0.09, 0.55}
\usepackage[breaklinks=true,bookmarks=false,colorlinks,urlcolor=rose]{hyperref}

\cvprfinalcopy % *** Uncomment this line for the final submission

\def\cvprPaperID{1006} % *** Enter the CVPR Paper ID here

\newcolumntype{Y}{>{\centering\arraybackslash}X}
\ifcvprfinal\fi
\begin{document}

%%%%%%%%% TITLE
\title{DAVANet: Stereo Deblurring with View Aggregation}
\author{Shangchen Zhou$^1$\ \ \ Jiawei Zhang$^1$\ \ \ Wangmeng Zuo$^{2}$\thanks{Corresponding author}\ \ \ \ Haozhe Xie$^2$\ \ \ Jinshan Pan$^3$\ \ \ Jimmy Ren$^1$\\
	$^1$SenseTime Research\quad
	$^2$Harbin Institute of Technology, Harbin, China\\
	$^3$Nanjing University of Science and Technology, Nanjing, China\\
	% {\tt\small \{zhoushangchen,zhangjiawei,rensijie\}@sensetime.com}\\
	{\tt\small \url{https://shangchenzhou.com/projects/davanet}}
}
\maketitle
\thispagestyle{empty}
%ABSTRACT
\begin{abstract}
	\vspace{-1.8mm}
	% ===============
	Nowadays stereo cameras are more commonly adopted in emerging devices such as dual-lens smartphones and unmanned aerial vehicles. 
	However, they also suffer from blurry images in dynamic scenes which leads to visual discomfort and hampers further image processing.
	Previous works have succeeded in monocular deblurring, yet there are few studies on deblurring for stereoscopic images. 
	By exploiting the two-view nature of stereo images, we propose a novel stereo image deblurring network with \textbf{D}epth \textbf{A}wareness and \textbf{V}iew \textbf{A}ggregation, named \textbf{DAVANet}. 
	In our proposed network, 3D scene cues from the depth and varying information from two views are incorporated, which help to remove complex spatially-varying blur in dynamic scenes.
	Specifically, with our proposed fusion network, we integrate the bidirectional disparities estimation and deblurring into a unified framework. 
	Moreover, we present a large-scale multi-scene dataset for stereo deblurring, containing 20,637 blurry-sharp stereo image pairs from 135 diverse sequences and their corresponding bidirectional disparities. 
	The experimental results on our dataset demonstrate that DAVANet outperforms state-of-the-art methods in terms of accuracy, speed, and model size.
	% ===============
\end{abstract}
\vspace{-2mm}
\section{Introduction}
%==============
\begin{figure}
	\centering
	\resizebox{0.96\linewidth}{!} {
		\includegraphics{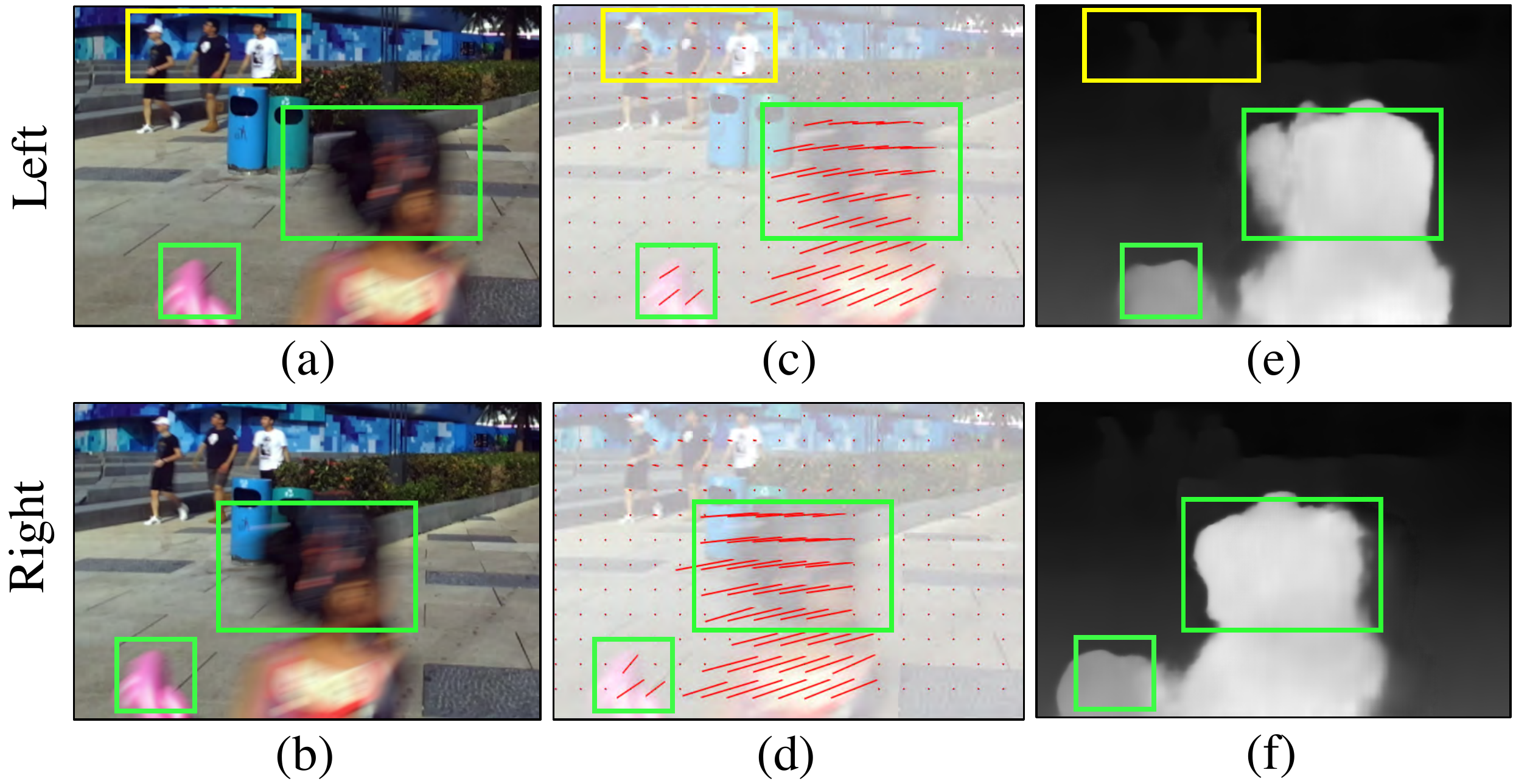}
	}
	\caption{Depth-varying and view-varying blur. (a, b) are the stereo blurry images, (c, d) are the motion trajectories in terms of optical flow which models the blur kernels and (e, f) are the estimated disparities. The objects with different depths have different blurs which can be seen between green and yellow boxes. In addition, the green boxes show that the blurs are different between two views. The proposed \textit{DAVANet} makes use of the above properties for deblurring.}
	\label{fig:varyingblur}
	\vspace{-2mm}
\end{figure}
% ================
With the wide use of dual-lens smartphones, unmanned aerial vehicles and autonomous robots, 
stereoscopic vision has attracted increasing attention from researchers.
Relevant studies not only covers traditional stereo tasks, such as stereo matching~\cite{lecun2015stereo, chang2018pyramid, pang2018zoom} and scene flow estimation~\cite{mayer2016large, moritz2015kitti,ilg2018occlusions},  
but also some novel tasks for improving visual effects of stereoscopic 3D contents, for example, stereo super-resolution~\cite{jeon2018enhancing}, stereo video retargeting~\cite{li2018depth} and stereo neural style transfer~\cite{chen2018stereoscopic, gong2018neural}. 
However, stereo image deblurring has rarely been discussed. In fact, the images captured by handheld or on-board stereo cameras often contain blur due to camera shake and object motion.
The blurry stereo images would cause visual discomfort to viewers and make it difficult for further image processing. 
% ===============

% ===============
Dynamic scene deblurring from a single blurry image is a highly ill-posed task. 
Due to depth variation and object/camera motion in dynamic scenes, it is difficult to estimate spatially variant blur with the limited information from single observation.
Although the existing CNN based methods~\cite{tao2018scale, zhang2018dynamic, kupyn2018deblurgan, aittala2018burst, nah2017deep, su2017deep} have achieved encouraging results in monocular deblurring,
they still fail when handling complicated non-uniform blur.
To the best of our knowledge, there are few traditional methods~\cite{xu2012depth, sellent2016stereo, pan2017simultaneous} proposed to exploit stereo information for deblurring, 
where a coarse depth or piecewise rigid 3D scene flow is utilized to estimate blur kernels in a hierarchical or iterative framework. 
However, they are time-consuming due to the complex optimization process.
%==============

%==============
With a stereopsis configuration, our motivation is based on two observations:
(\romannumeral1) Depth information can provide helpful prior information for estimating spatially-varying blur kernels. The near points are more blurry than the distant ones in a static scene which can be seen between the green and yellow boxes in Figure~\ref{fig:varyingblur}.  
Compared monocular-based algorithms, the proposed stereo-based method can obtain more accurate depth information by the disparity estimation.
(\romannumeral2) The varying information in corresponding pixels cross two stereo views can help blur removal.
In Section~\ref{sec:motivation}, we demonstrate that the corresponding pixels in two different views have different blurs due to the motion perpendicular towards the camera and rotation, which is shown as the green boxes in Figure~\ref{fig:varyingblur}.
The network can benefit from aggregated information, where the sharper pixel can be transferred and selected by using an adaptive fusion scheme.
Two views can also share varying information, e.g., non-occlusion areas, caused by different viewpoints.
% ==============

% ==============
Inspired by these two insights, we propose a novel depth-aware and view-aggregated stereo deblurring network, named \textit{DAVANet}. 
It consists of \textit{DeblurNet} and \textit{DispBiNet}, for image deblurring and bidirectional disparities estimation respectively.
The \textit{DeblurNet} and the \textit{DispBiNet} are integrated at feature domain by the proposed fusion network, named \textit{FusionNet}. 
Specifically, the \textit{DispBiNet} provides depth-integrated features and bidirectional disparities for the \textit{FusionNet}.
The \textit{FusionNet} fully exploits these inputs and enriches the \textit{DeblurNet} features with embedding depth and the other view information. 
With the perception of 3D scene information from stereo images, the proposed method is effective for dynamic scene deblurring.
Finally, to obtain richer contextual information, a context module is designed to incorporate the multi-scale contextual information by applying several parallel atrous convolutions with different dilation rates.
% ==============

% ==============
Currently, there is no particular dataset for stereo deblurring.
As a result, we propose a large-scale multi-scene stereo blurry image dataset.
It consists of 20,637 blurry-sharp stereo image pairs from 135 different sequences (98 for training and 37 for testing) and corresponding bidirectional disparities obtained from the ZED stereo camera~\cite{stereolabs}. 
We adopt the blur generation method used in~\cite{li2010generating, nah2017deep, su2017deep}, that is, approximating a longer exposure by accumulating the frames in an image sequence.
We first interpolate frame of captured videos to a very high frame rate (480 fps) using frame interpolation method proposed in~\cite{niklaus2017iccv}  and then average the sharp sequence to create a blurry image.
% ==============

% ==============
The main contributions are summarized as follows:
\vspace{-3mm}
\begin{itemize}
	\setlength{\itemsep}{1pt}
	\setlength{\parsep}{0pt}
	\setlength{\parskip}{1pt}
	\item We propose a unified network for stereo deblurring. The \textit{DispBiNet} predicts the bidirectional disparities for depth awareness as well as view information aggregation in the \textit{FusionNet}, which helps the \textit{DeblurNet} to remove dynamic scene blur from stereo images.
	\item We present a first large-scale multi-scene dataset for stereo deblurring, which consists of 20,637 stereo images from 135 diverse scenes. It is currently the largest dataset for deblurring.
	\item We both quantitatively and qualitatively evaluate our method on our dataset and show that it performs favorably against state-of-the-art algorithms in terms of accuracy, speed as well as model size.
\end{itemize}

\section{Related Work}
%==================
Our work is a new attempt for solving stereo image deblurring by integrating blur removal and disparity estimation into a unified network.
The following is a review of relevant works on monocular single-image deblurring, monocular multi-image deblurring, as well as stereo image deblurring respectively.
% =================

% =================
\noindent \textbf{Single-image Deblurring.} 
Many methods have been proposed for single-image deblurring. Some natural image priors are designed to help blur removal, such as $L_0$-regularized prior~\cite{xu2013unnatural}, dark channel prior~\cite{pan2016blind}, and discriminative prior~\cite{li2019blind}.
However, it is difficult for these methods to model spatially variant blur in dynamic scenes. 
To model the non-uniform blur, some depth-based methods~\cite{lee2017joint, park2017joint, hu2014joint, paramanand2013non} that utilize the predicted depth map to estimate different blur kernels.
When the blur kernels are not be accurately estimated, they tend to generate visual artifacts in restored images. 
Moreover, they are computationally inefficient due to the complex optimization process. 
% ================

% ================
Recent years have witnessed significant advances in single image deblurring by CNN-based models.
Several methods~\cite{sun2015learning, gong2017motion}  use CNNs to estimate the non-uniform blur kernels. A conventional non-blind deblurring algorithm~\cite{zoran2011learning} is used removing blur, which is time-consuming.
More recently, many end-to-end CNN models for image deblurring have also been proposed~\cite{nah2017deep, noroozi2017motion, zhang2017learning, tao2018scale, zhang2018dynamic, kupyn2018deblurgan}. To obtain a large receptive field in the network for blur removal, \cite{tao2018scale} and \cite{tao2018scale} develop a very deep multi-scale networks in coarse-to-fine manner. 
Different from~\cite{nah2017deep}, Tao \textit{et al.}~\cite{tao2018scale} share the weights of the network at three different spatial scales and use the LSTM to propagate information across scales. 
To handle spatially variant blur in dynamic scenes, Zhang \textit{et al.}~\cite{zhang2018dynamic} adopt a VGG network to estimate the pixel-wise weights of the spatially variant RNNs~\cite{liu2016learning} for blur removal in feature space. 
Noroozi \textit{et al.}~\cite{noroozi2017motion} build skip connections between the input and output, which reduces the difficulty of restoration and ensures color consistency.
In addition, the adversarial loss is used in~\cite{nah2017deep, kupyn2018deblurgan} to restore more texture details.
% =================

% =================
\noindent \textbf{Multi-image Deblurring.} 
Recently, several CNN-based methods~\cite{su2017deep, hyun2017online, kim2018spatio, aittala2018burst} have been proposed for monocular multi-image (video/burst) deblurring.
\cite{su2017deep} and \cite{kim2018spatio} align the nearby frames with the reference frame to restore the sharp images, which can obtain more rich information cross different images.
Kim \textit{et al.}~\cite{hyun2017online} propose a frame recurrent network to aggregate multi-frame features for video deblurring.
By repeatedly exchanging the features across the burst images, Aittala \textit{et al.}~\cite{aittala2018burst} propose an end-to-end burst deblurring network in an order-independent manner.
Based on the observations that the different images from video or burst are blurred differently, these multi-image fusion methods usually lead to good performance.
%
% =================

% =================
\noindent \textbf{Stereo Deblurring.} 
So far, there are few traditional methods~\cite{xu2012depth, sellent2016stereo, pan2017simultaneous} that leverage the scene information (i.e., disparity and flow) from stereo images for deblurring.
Xu and Jia~\cite{xu2012depth} partition the image into regions according to disparity (depth) estimated from stereo blurry images and estimate their blur kernels hierarchically.
The methods~\cite{sellent2016stereo, pan2017simultaneous} propose a stereo video deblurring framework, where  3D scene flow estimation and blur removal are conducted jointly so that they can enhance each other with an iterative manner.
%
% =================
\section{Proposed Method}
\subsection{Motivation}
\label{sec:motivation}
The motivation that utilizing stereo camera for dynamic scene deblurring is inspired by two observations, which is exemplified in Figure~\ref{fig:varyingblur}.
First, we find that nearby object points are more blurry than distant ones and stereo cameras can provide depth information (disparity).
Second, the two views of the stereo camera may produce different sizes of the blur to the same object because of relative motion along the depth direction and camera rotation.
The sharper view can help the other view to restore better by sharing its information. 
In this section, we analyze the above observations in details with the assumption that the stereo camera has already been rectified.
% =================

\begin{figure}
	\centering
	\resizebox{\linewidth}{!} {
		\includegraphics{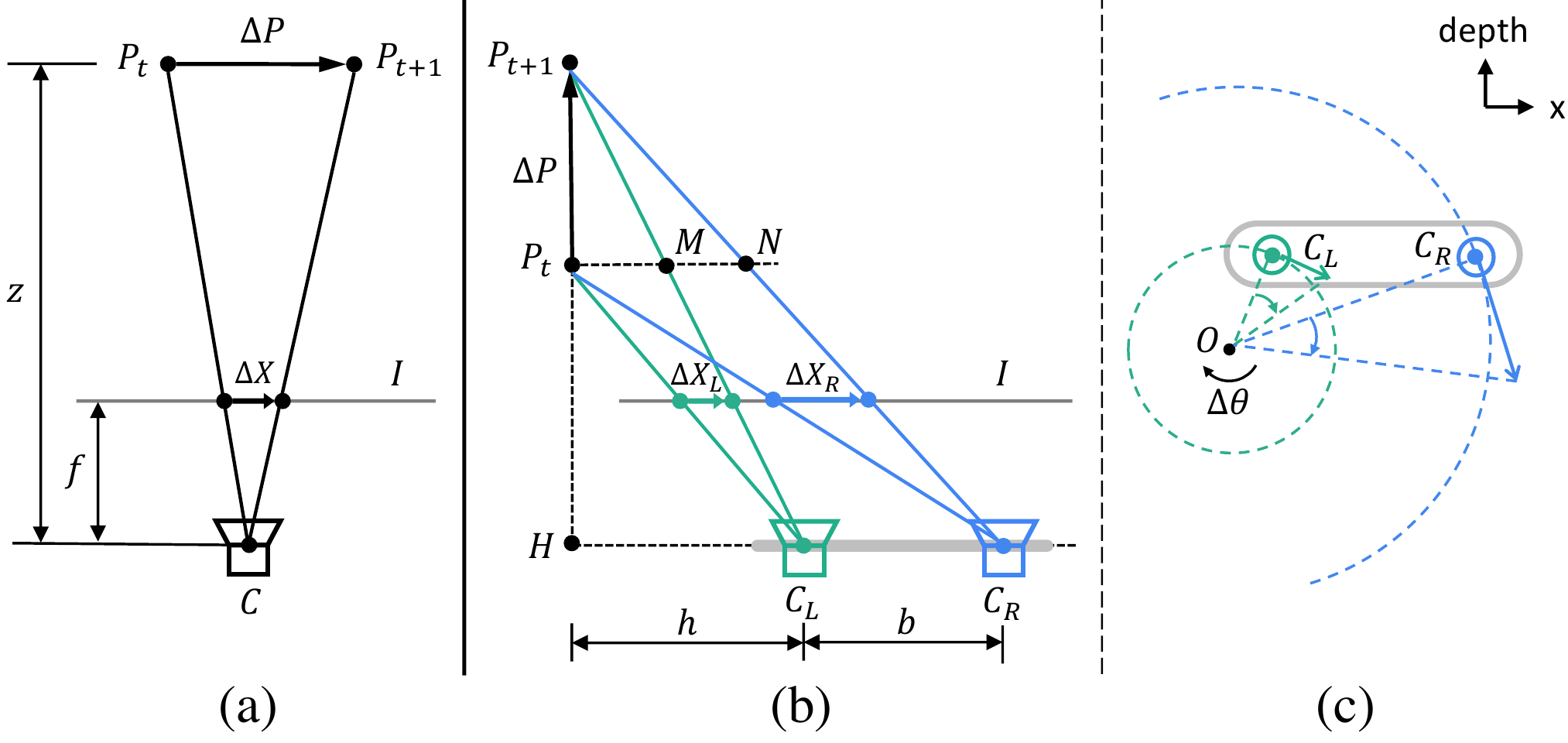}
	}
	\caption{(a) is the depth-varying blur due to relative translation parallel to the image plane.
		(b) and (c) are the view-varying blur due to relative translation along depth direction and rotation.
		Note that all complex motion can be divided into above three relative sub-motion patterns.}
	\label{fig:depth_view_blur}
	\vspace{-3mm}
\end{figure}

% =================
\noindent \textbf{Depth-Varying Blur.}
In \cite{xu2012depth}, Xu and Jia have analyzed the relationship between blur size and depth. In Figure~\ref{fig:depth_view_blur}(a), we simply restate it by only considering the relative translation parallel to the image plane $I$.
According to the similar triangles theorem:
\begin{equation}
\label{eq:depth_blur}
{\Delta X}/{\Delta P} = {f}/{z},
\end{equation}
in which $\Delta X$, $\Delta P$, $f$ and $z$ denote the size of blur, the motion of object point, focal length, and depth of object point, respectively.
Eq.~\ref{eq:depth_blur} shows that blur size $\Delta X$ is inversely proportional to depth $z$ if motion $\Delta P$ is fixed, which means that the closer object will generate the larger blur.
% =================

% =================
\noindent \textbf{View-Varying Blur.}
For the stereo setups, the relative movements between the object point $P$ and two lens of stereo camera are different because the point $P$ is captured from different viewpoints.
These differences make the object exhibit different blurs under the two views.
Here, we consider two scenarios: relative translation along depth direction and rotation.
For translation, we assume the object point $P$ moves from $P_t$ to $P_{t+1}$ along the depth direction in Figure~\ref{fig:depth_view_blur}(b).
According to the similar triangles theorem:
\begin{equation}
\label{eq:view_blur_trans}
{\Delta X_L}/{\Delta X_R}={\overline{P_tM}}/{\overline{P_tN}}={h}/{(h+b)},
\end{equation}
where $b$ is the baseline of the stereo camera and $h$ is the distance between left camera $C_L$ and line $\overline{P_tP_{t+1}}$.
It demonstrates that the blur sizes for two views of a stereo camera are different due to relative translation in depth direction. 
% =================

% =================
As to relative rotation in Figure~\ref{fig:depth_view_blur}(c), the velocities of two lens $v_{C_L}, v_{C_R}$ of the stereo camera are proportional to the corresponding radiuses of the rotation $\overline{C_LO}$, $\overline{C_RO}$:
\begin{equation}
\label{eq:view_blur_rot}
{v_{C_L}}/{v_{C_R}}={\overline{C_LO}}/{\overline{C_RO}}.
\end{equation}
In addition, the directions of the velocities are different due to relative rotation.
As a result, both the size and direction of the blur vary between two views.
The proposed network can utilize the information from the clearer view to help restore a better image for the more blurry one.
% =================
% =================
% ==============
\begin{figure*}
	\centering
	\resizebox{0.85\linewidth}{!} {
		\includegraphics{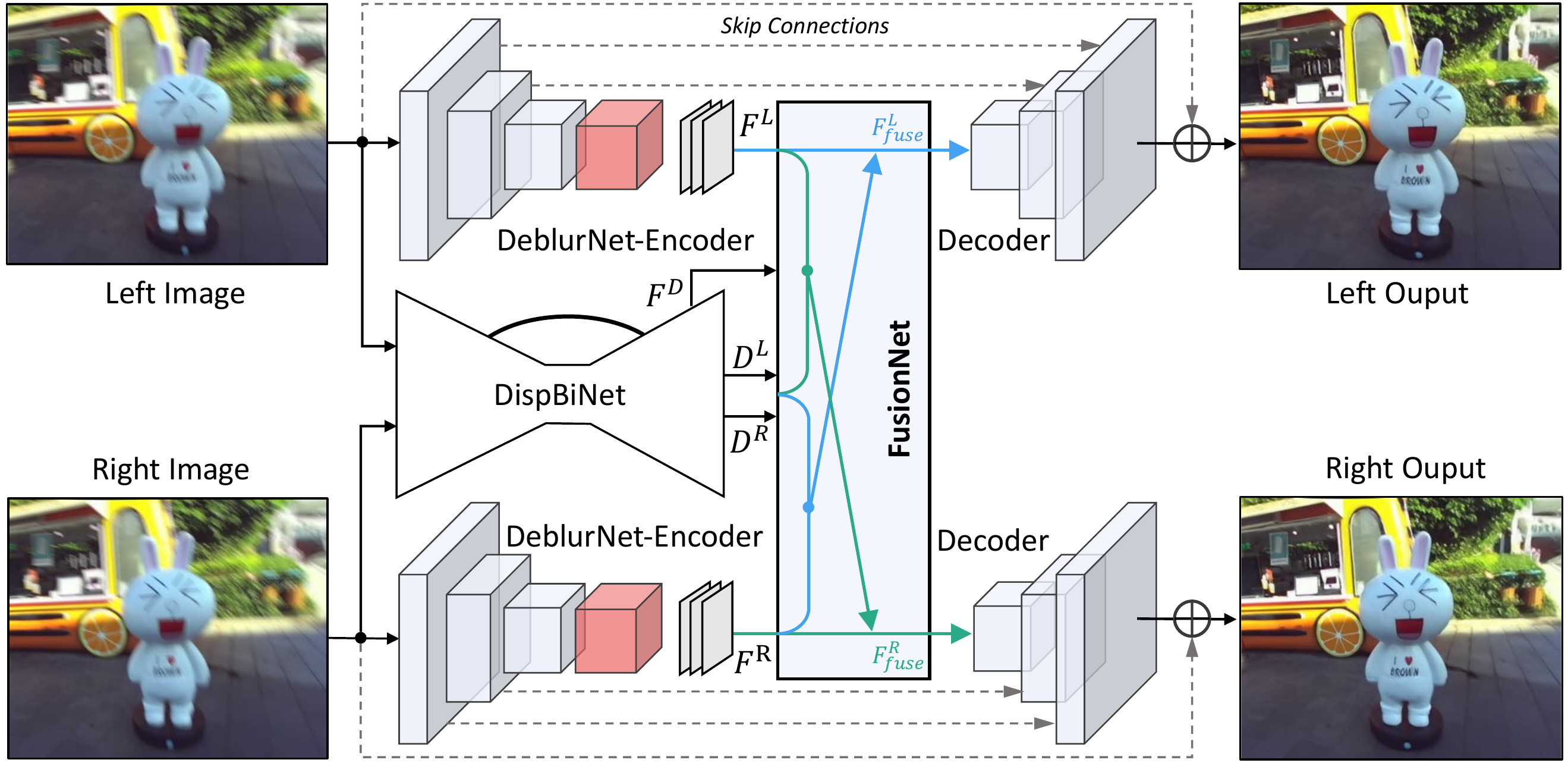}
	}
	\caption{The overall structure of stereo deblurring network \textit{DAVANet}, where the depth and the two-view information from the \textit{DispBiNet} and the \textit{DeblurNet} are integrated in \textit{FusionNet}. Note that the \textit{DeblurNet} shares weights for two views.}
	\vspace{-1mm}
	\label{fig:pipeline}
\end{figure*}
% ==============
\subsection{Network Architecture}
% ==============
The overall pipeline of the proposed \textit{DAVANet} is illustrated in Figure~\ref{fig:pipeline}. 
It consists of three sub-networks:  \textit{DeblurNet} for single-image deblurring, \textit{DispBiNet} for bidirectional disparities estimation and \textit{FusionNet} for fusing depth and two-view informations in an adaptive selection manner.
Note that we adopt small convolution filters $(3\times3)$ to construct these three sub-networks and find that using the large filters does not significantly improve the performance.
% ==============

\begin{figure*}
	\centering
	\resizebox{\linewidth}{!} {
		\includegraphics{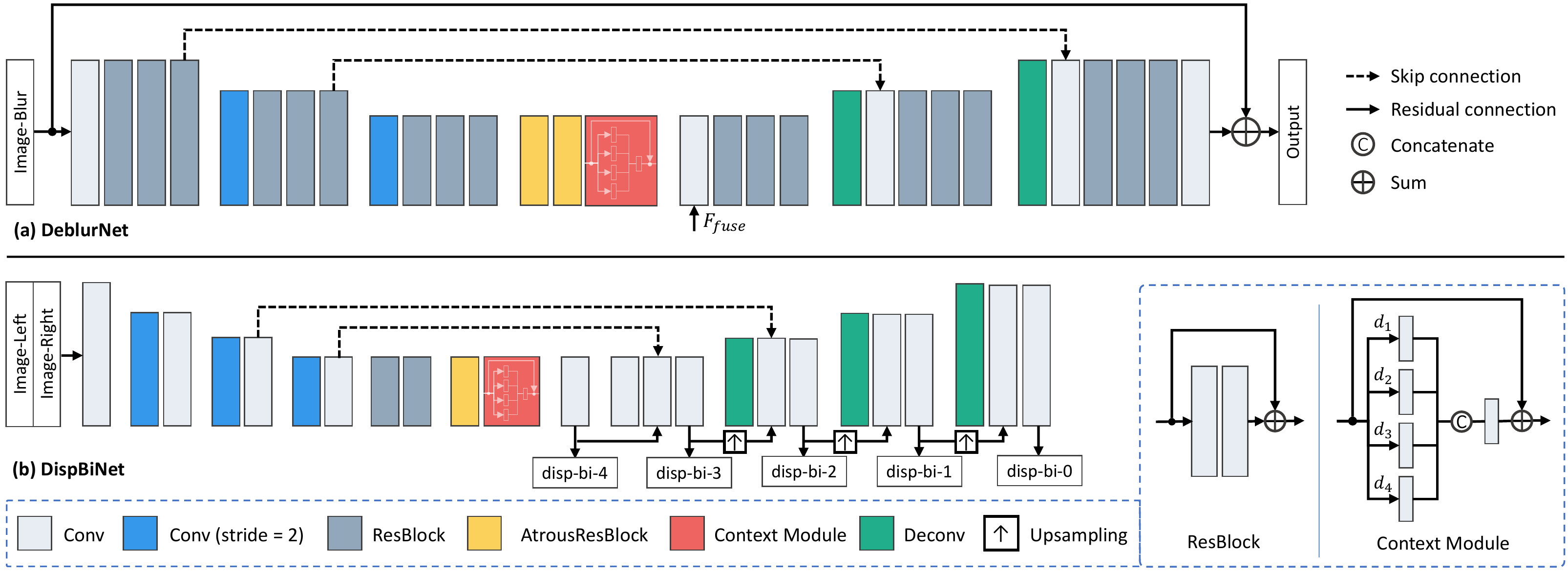}
	}
	\caption{The detail structures of \textit{DeblurNet} and \textit{DispBiNet}. To get richer multi-scale features, the \textit{Context Module} is adopted in both \textit{DeblurNet} and \textit{DispBiNet}, which contains parallel dilated convolutional layers with different dilation rates.}
	\vspace{-1mm}
	\label{fig:networks}
\end{figure*}

% ==============
\noindent \textbf{DeblurNet.}
The U-Net based structure of \textit{DeblurNet} is shown in Figure~\ref{fig:networks}(a). 
We use the basic residual block as the building block, which has been proved effectiveness in deblurring~\cite{nah2017deep, tao2018scale}.
The encoder outputs features with $\frac{1}{4}\times\frac{1}{4}$ of the input size.
Afterward, the following decoder reconstructs the sharp image with full resolution via two upsampled residual blocks.
The skip-connections between corresponding feature maps are used between encoder and decoder.
In addition, we also adopt a residual connection between the input and output.
which makes it easy for the network to estimate the residual between blurry-sharp image pair and maintains color consistency.
% =============

% =============
To enlarge the receptive field and obtain the multi-scale information, the scale-recurrent scheme is popularly adopted in~\cite{nah2017deep, tao2018scale}.
Despite their performance improvement, they greatly increase the complexity of time and space.
To solve this, we employ the two atrous residual blocks and a \textit{Context Module} between encoder and decoder to obtain richer features.
The \textit{Context module} will be described in later a section. It should be noted that the \textit{DeblurNet} uses shared weights for both views. 
% =============

% =============
\noindent \textbf{DispBiNet.}
Inspired by DispNet~\cite{mayer2016large} structure, we propose a small \textit{DispBiNet} as shown in Figure~\ref{fig:networks}(b).
Different from DispNet, the proposed \textit{DispBiNet} can predict bidirectional disparities in one forward process.
The bidirectional prediction has been proved better than unidirectional prediction in scene flow estimation~\cite{ilg2018occlusions}.
The output is the full resolution with three times downsample and upsample in this network.
In addtion, the residual block, atrous residual block, and context module are also used in \textit{DispBiNet}.
% =============

% =============
\noindent \textbf{Context Module.}
To embed the multi-scale features, we propose the \textit{Context Module} (a slightly modified version of ASPP~\cite{chen2018deeplab}) for \textit{DeblurNet} and \textit{DispBiNet}, which contains parallel dilated convolutions with different dilated rates, as show in Figure~\ref{fig:networks}.
The four dilated rates are set to: $1, 2, 3, 4$.
\textit{Context Module} fuses richer hierarchical context information that benefit both blur removal and disparity estimation.
% =============

\begin{figure}
	\centering
	\resizebox{0.98\linewidth}{!} {
		\includegraphics{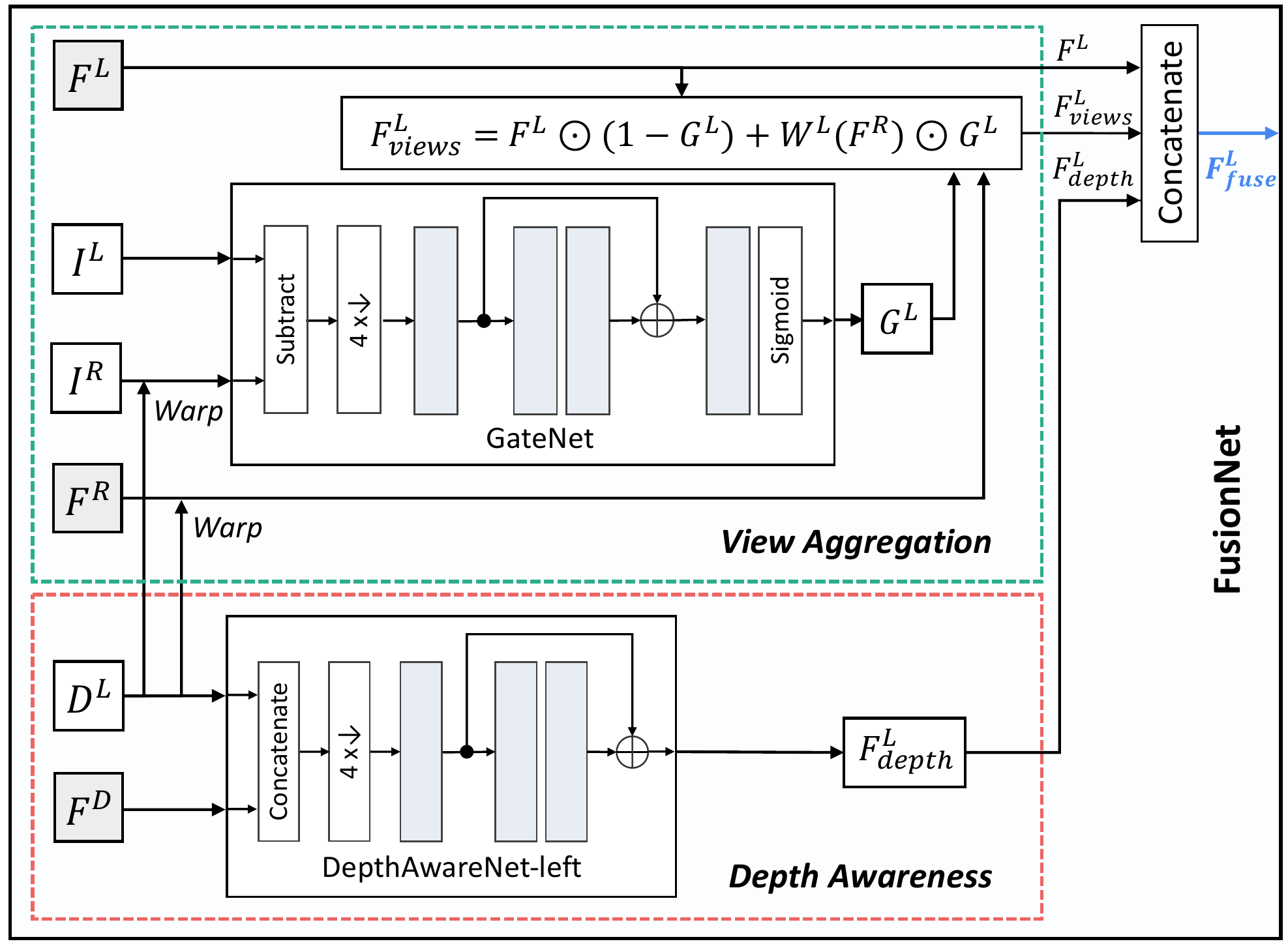}
	}
	\caption{Fusion network. The \textit{FusionNet} consists two components: depth awareness and view aggregation, which generate the depth-view fused feature for the decoder of \textit{DeblurNet}. For simplicity, we only show the forward pass for the left image.}
	\vspace{-3mm}
	\label{fig:fusion}
\end{figure}

% =============
\noindent \textbf{Fusion Network.}
To exploit depth and two-view information for deblurring, we introduce the fusion network \textit{FusionNet} to enrich the features with the disparities and the two views.
For simplicity, we take left image as reference in this sections. 
As shown in Figure~\ref{fig:fusion}, \textit{FusionNet} takes the original stereo images $I^L, I^R$, the estimated disparity of left view $D^L$, features $F^D$ of the second last layer of \textit{DispBiNet} and features $F^L, F^R$ from \textit{DeblurNet} encoder as input in order to generate the fused features $F_{fuse}^L$.
% =============

% =============
For two-view aggregation, the estimated left-view disparity $D^L$ is used to warp right-view features $F^R$ of \textit{DeblurNet} to the left view, denoted as $W^L(F^R)$. 
Instead of directly concatenating $W^L(F^R)$ and $F^L$, the sub-network \textit{GateNet} is employed to generate a soft gate map $G^L$ ranging from 0 to 1.
The gate map can be utilized to fuse features $F^L$ and $W^L(F^R)$ in an adaptive selection scheme, that is, it selects helpful features and rejects incorrect ones from the other view.
For example, at occlusion or false disparity regions, the values in the gate map tend to be 0, which suggest that only the features of reference view $F^L$ should be adopted.
The \textit{GateNet} consists of five convolutional layers as shown in Figure~\ref{fig:fusion}. Its input is absolute difference of input left image $I^L$ and the warped right image $W^L(I^R)$, namely $\left| I^L - W^L(I^R)\right|$,
and the output is a single channel gate map. All feature channels share the same gate map to generate the aggregated features:
\begin{equation}
F_{views}^L = F^L\odot(1-G^L)+W^L(F^R)\odot G^L,
\end{equation}
where $\odot$ denotes element-wise multiplication.
% =============
% =============

% =============
For depth awareness, a sub-network \textit{DepthAwareNet} containing three convolutional layers is employed,
and note that this sub-network is not shared by both views.
Given the disparity $D^L$ and the second last layer features $F^D$ of \textit{DispBiNet}, \textit{DepthAwareNet-left} produces the depth-involved features $F_{depth}^L$.
In fact, \textit{DepthAwareNet} learns the depth-aware prior implicitly, which helps for dynamic scene blur removal.
% =============

% =============
Finally, we concatenate the original left-view features $F^L$, view-aggregated features $F_{views}^L$, and depth-aware features $F_{depth}^L$ to generate the fused left-view features $F_{fuse}^L$. 
And then, we feed the $F_{fuse}^L$ to the decoder of \textit{DeblurNet}. Note that the fusion processings of two views are the same.
\subsection{Losses}
\noindent \textbf{Deblurring Losses.}
For Deblurring, we consider two loss functions to measure the difference between the restored image $\hat{I}$ and sharp image $I$ for both two views $L, R$. The first loss is MSE loss:
\begin{equation}
\label{eq:mse_loss}
\mathcal{L}_{mse} = \frac{1}{2CHW}\sum _{k \in \{L, R\}}  || \hat{I}^k- I^k ||^2,
\end{equation}
where $C, H, W$ are dimensions of image. The other loss function is perceptual loss proposed in~\cite{johnson2016perceptual}, which is defined as the $l_2$-norm between the VGG-19~\cite{simonyan2015very} features of restored image $\hat{I}$ and sharp image $I$:
\begin{equation}
\label{eq:percept_loss}
\mathcal{L}_{perceptual} = \frac{1}{2\mathcal{C}_j\mathcal{H}_j\mathcal{W}_j}\sum _{k \in \{L, R\}}  ||\Phi_j(\hat{I}^k) - \Phi_j(I^k)||^2,
\end{equation}
where $\mathcal{C}_j, \mathcal{H}_j, \mathcal{W}_j$ are dimensions of the features, and $\Phi_j(\cdot)$ denotes the features from the $j$-th convolution layer within the pretrained VGG-19 network. In our work we use the features from conv3-3 layer ($j$=15).
The overall loss function for deblurring is:
\begin{equation}
\label{eq:deblur_loss}
\mathcal{L}_{deblur} = \sum \limits_{k \in \{L, R\}}  w_1 \mathcal{L}_{mse}^k + w_2 \mathcal{L}_{perceptual}^k,
\end{equation}
where the weights $w_1, w_2$ of two losses are set to $1, 0.01$ in our experiments, respectively.
% =============

% =============
\noindent \textbf{Disparity Estimation Loss.}
For training \textit{DispBiNet}, we consider MSE loss between estimated disparities $\hat{D}$ and ground truth $D$ at multiple scales and remove the invalid and occlusion regions with mask map $M$:
\begin{equation}
\label{eq:disp_loss}
\mathcal{L}_{disp} = \sum \limits_{k \in \{L, R\}} \sum \limits_{i = 1} ^ m \frac{1}{H_iW_i} ||\hat{D}_i^k - D_i^k||^2\odot M_i^k,
\end{equation}
where $m$ is the number of scales of the network and the loss at each scale $i$ is normalized. 
% =============
\section{Stereo Blur Dataset}
\label{sec:data}
Currently, there is no dataset specially designed for stereo image deblurring. Therefore, to train our network and verify its effectiveness, we propose a large-scale, multi-scene and depth-varying stereo blur dataset.
It consists of a wide variety of scenarios, both indoor and outdoor. The indoor scenarios collect objects and persons, which usually with small depth. 
The outdoor scenarios include pedestrians, moving traffic and boats as well as natural landscapes. 
Moreover, we have diversified the dataset by considering various factors including illumination and weather. 
In the meantime, we have different photograph fashions including handheld shots, fixed shots, and onboard shots, to cover diverse motion patterns.
% ============

% ============
Inspired by the dynamic scene blur image generation method in~\cite{nah2017deep, su2017deep, hirsch2011fast}, 
we average a sharp high frame rate sequence to generate a blurry image to approximate a long exposure. 
In practice, we use the ZED stereo camera~\cite{stereolabs} to capture our data, which has the highest frame rate (60 fps) among the available stereo cameras.
However, the frame rate is still not high enough to synthesize look-realistic blur, without generating undesired artifacts which exist in GOPRO dataset~\cite{nah2017deep}. 
Therefore, we increase the video frame rate to 480 fps using a fast and high-quality frame interpolation method proposed in~\cite{niklaus2017iccv}. 
Then, we average the varying number (17, 33, 49) of successive frames to generate different blur in size,
which is temporally centered on a real-captured sharp frame (ground truth frame).
For the synthesis, both two views of the stereo video have the same settings.
In addition, to explore how the depth information helps with deblurring, our dataset also provides the corresponding bidirectional disparity of two views, acquired from a ZED camera. 
We also present the mask map for removing the invalid values in disparity ground truth and occlusion regions obtained by bidirectional consistency check~\cite{sundaram2010dense}.
% ============

% ============
In total, we collect 135 diverse real-world sequences of dynamic scenes.
The dataset consists of 20,637 blurry-sharp stereo image pairs with their corresponding bidirectional disparities at $1280 \times 720$ resolution.
We divide the dataset into 98 training sequences (17,319 samples) and 37 testing sequences (3,318 samples).
The scenarios are totally different for training and testing sets, which avoids the over-fitting problem.
\section{Experiments}
\subsection{Implementation Details}
In our experiments, we train the proposed single and stereo image deblurring networks (i.e., \textit{DeblurNet} and \textit{DAVANet}) using our presented Stereo Blur Dataset. 
For more convincing comparison with single-image methods, we also train and evaluate \textit{DeblurNet} on public GOPRO dataset~\cite{nah2017deep}, which contains 3,214 blurry-sharp image pairs (2,103 for training and 1,111 for evaluation). 
% ============

% ============
\noindent\textbf{Data Augmentation.}
Despite our large dataset, we perform several data augmentation techniques to add diversity into the training data.
We perform geometric transformations (randomly cropped  to $256\times 256$ patches and  randomly flipped  vertically) and chromatic transformations (brightness, contrast and saturation are uniformly sampled within $\left[0.8, 1.2\right]$) using ColorJitter in PyTorch. To make our network robust, a Gaussian random noise from $\mathcal N(0, 0.01)$ is added to the input images. To keep the epipolar constraint of stereo images, we do not adopt any rotation and horizontal flip for data augmentation.
% ============

% ============
\noindent\textbf{Training.}
\label{sec:training}
The overall proposed network \textit{DAVANet} contains three sub-networks: \textit{DeblurNet}, \textit{DispBiNet} and \textit{FusionNet}.
We first pretrain our \textit{DeblurNet} and \textit{DispBiNet} on each task separately, then add \textit{FusionNet} to the network and train them jointly as a whole.
For all models, we set batch size to 2 and use the Adam~\cite{kingma2015adam} optimizer with parameters $\beta_1 = 0.9$ and $\beta_2 = 0.999$. 
The initial learning rate in our experiments is set to $10^{-4}$ and decayed by 0.5 every 200k iterations. 
% ============

% ============
For the \textit{DeblurNet}, we first train it on the presented dataset, where 2,000k iterations are sufficient for convergence.
% ============
For the \textit{DispBiNet}, we first train it using a subset (10,806 samples) of \textit{FlyingThings3D} dataset. 
In this subset, the samples with large disparity ($>90$ $pixels$) are removed to ensure that the distribution of its disparity is the same as our dataset. 
Then we finetune the \textit{DispBiNet} fully on our Stereo Blur Dataset until convergence.
Finally, we jointly train the overall network on our dataset for 500k iterations.
% ============
\subsection{Experimental Results}
\begin{table*}
	\centering
	\caption{Quantitative evaluation on our Stereo Blur Dataset, in terms of PSNR, SSIM, running time and parameter number. All existing methods are evaluated using their publicly available code. A “-” indicates that the result is not available.  Note that the running time for our stereo deblurring network (\textit{DAVANet}) records the forward time of both left and right images.}
	\resizebox{\linewidth}{!} {
		\begin{tabular}{lccccccccc}
			\toprule
			Method
			& Whyte~\cite{whyte2012non}
			& Sun~\cite{sun2015learning}
			& Gong~\cite{gong2017motion}
			& Nah~\cite{nah2017deep}
			& Kupyn~\cite{kupyn2018deblurgan}
			& Zhang~\cite{zhang2018dynamic}
			& Tao~\cite{tao2018scale}
			& Ours-Single
			& Ours-Stereo\\
			\midrule
			% single psnr matlab 32.07; stereo psnr in matlab 33.13
			PSNR     & 24.84      & 26.13      & 26.51      & 30.35       & 27.81      & 30.46     & 31.65  &   \bf{31.97}    & \bf{33.19}\\
			SSIM      & 0.8410     & 0.8830      & 0.8902      & 0.9294    & 0.8895     & 0.9367      & 0.9479  &  \bf{0.9507}     &  \bf{0.9586}\\
			\midrule
			Time (sec)       & 700      & 1200      & 1500       & 4.78           & 0.22       & 1.40     & 2.52       & \bf{0.13}      & \bf{0.31 / pair}\\
			Params (M)    & -          & 7.26        & 10.29      & 11.71     & 11.38      & 9.22      & 8.06    & \bf{4.59}    & 8.68\\
			\bottomrule
		\end{tabular}
	}
	\label{tab:stereo_psnr_time_size}
	
\end{table*}
% ============

\begin{figure*}[t]\footnotesize
	\centering
	\renewcommand{\tabcolsep}{1pt}
	\renewcommand{\arraystretch}{1}
	\begin{center}
		\begin{tabular}{ccccc}
			\includegraphics[width=0.19\linewidth]{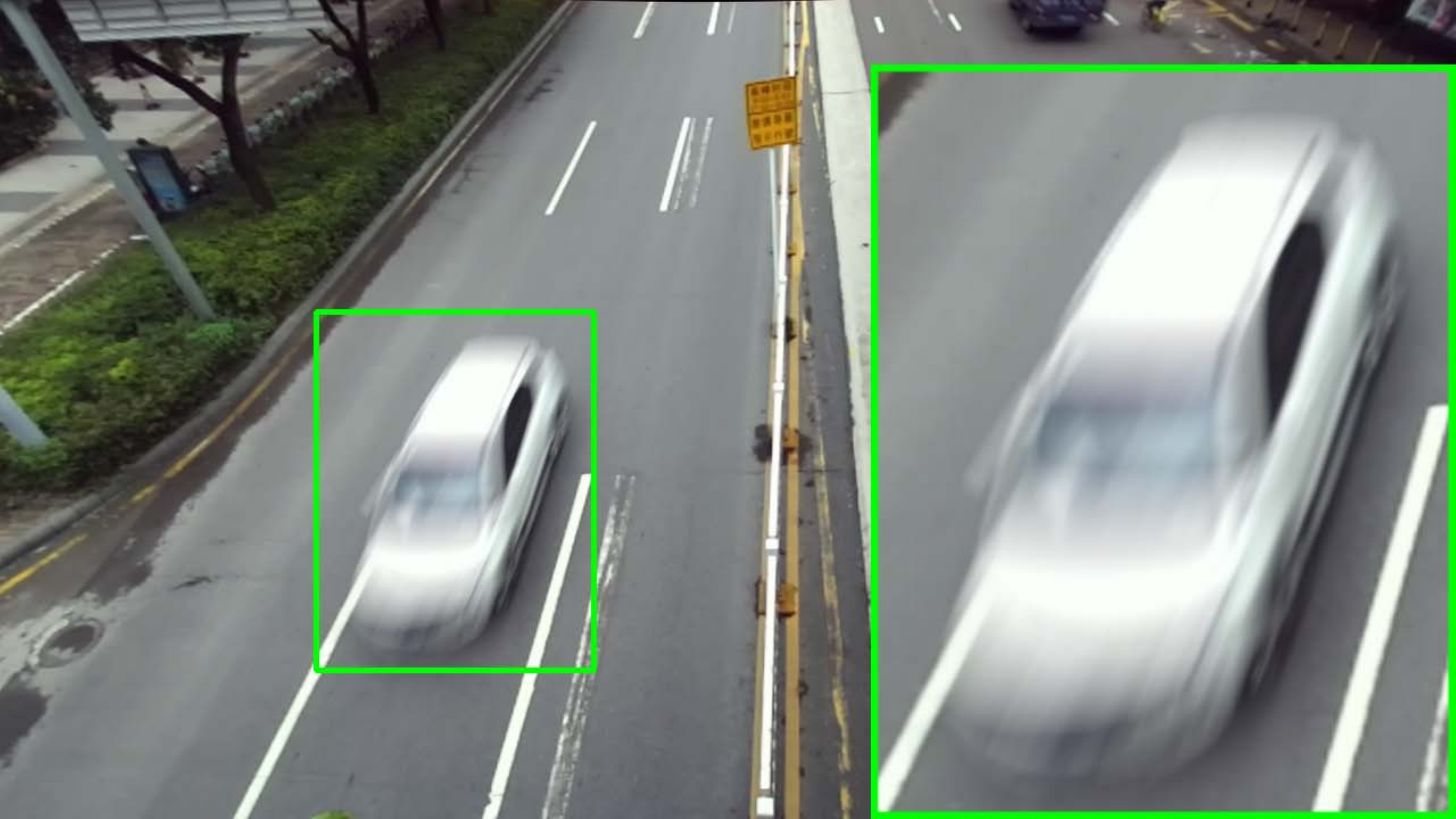} &
			\includegraphics[width=0.19\linewidth]{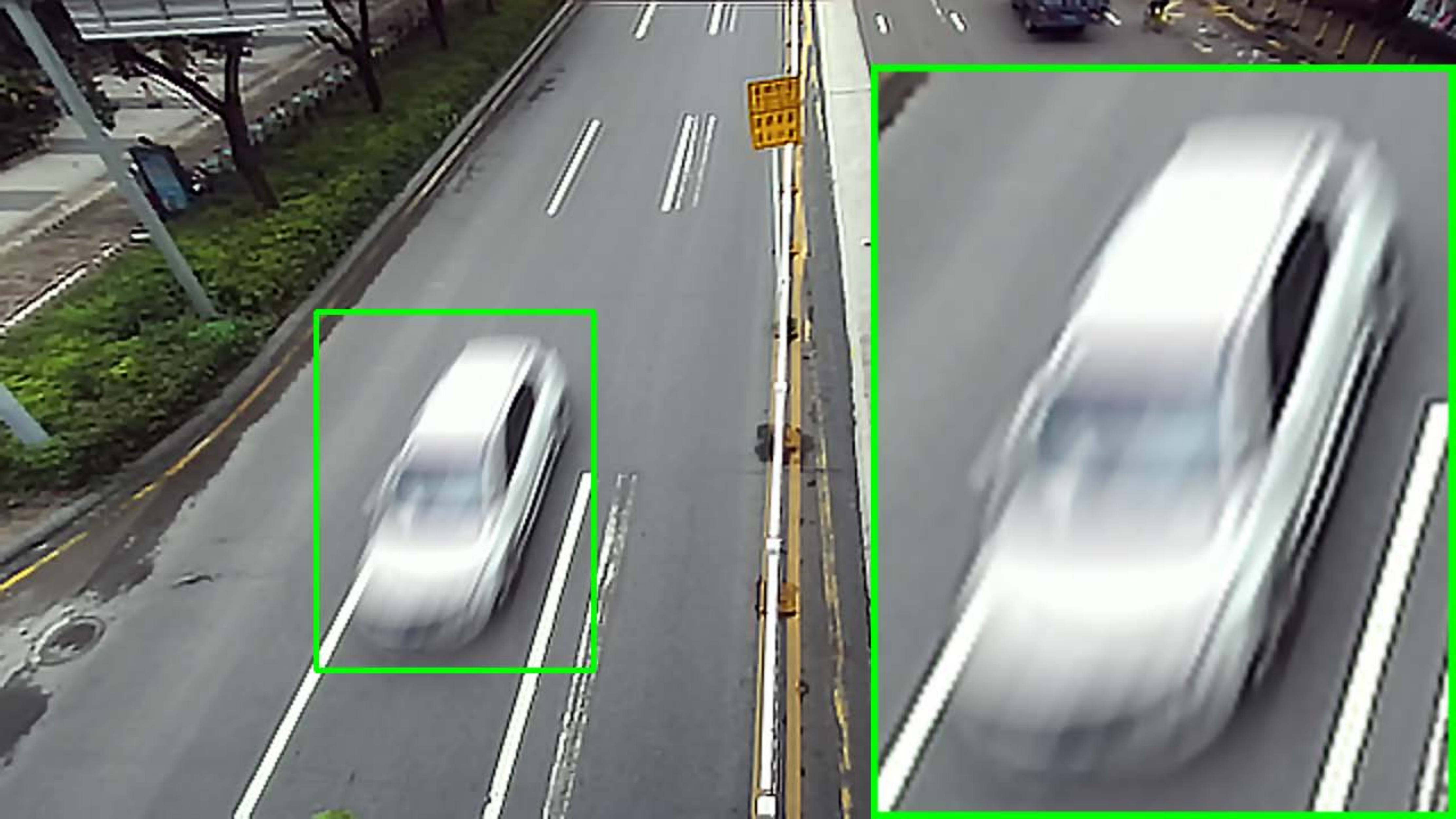} &
			\includegraphics[width=0.19\linewidth]{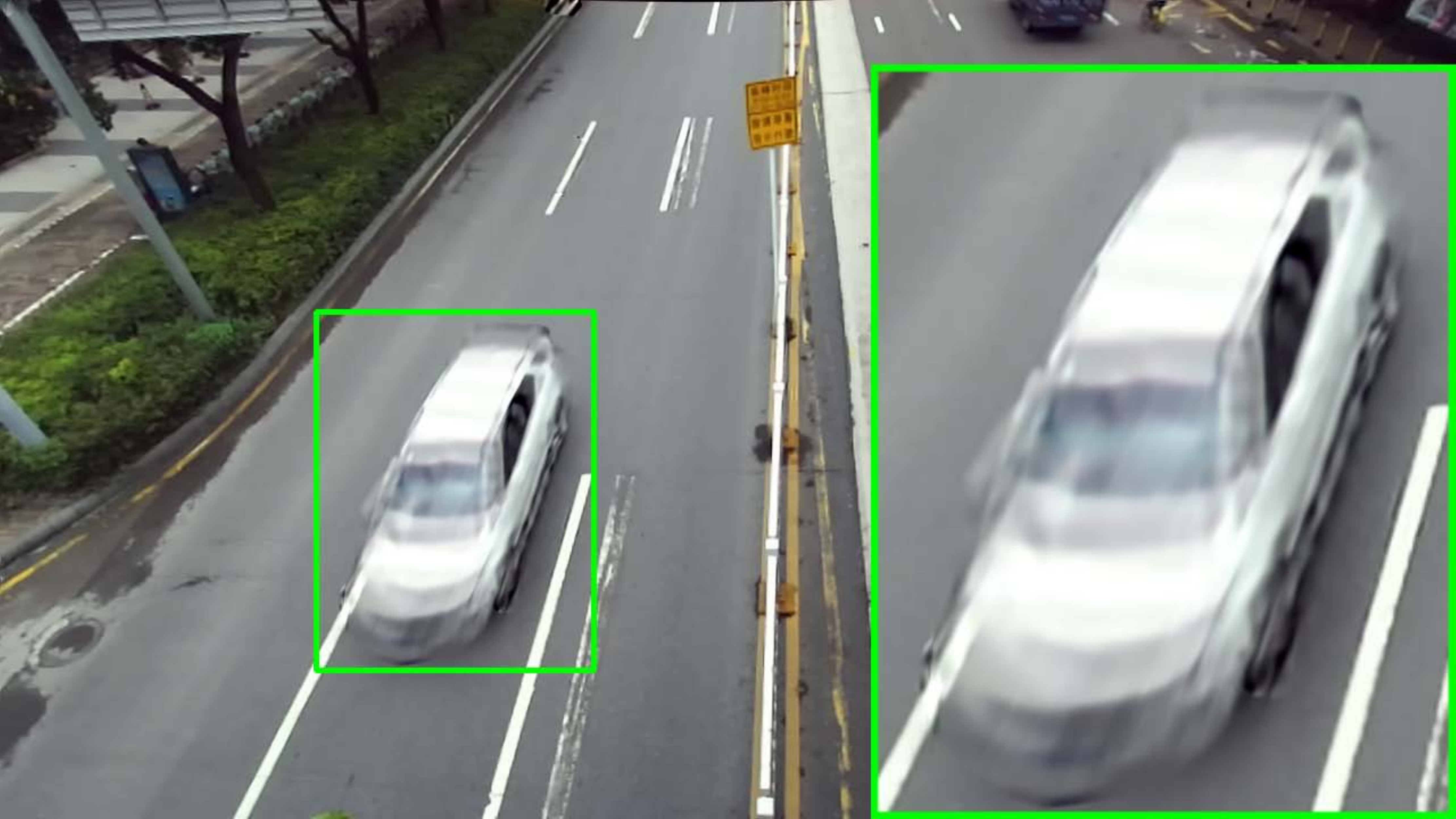} &
			\includegraphics[width=0.19\linewidth]{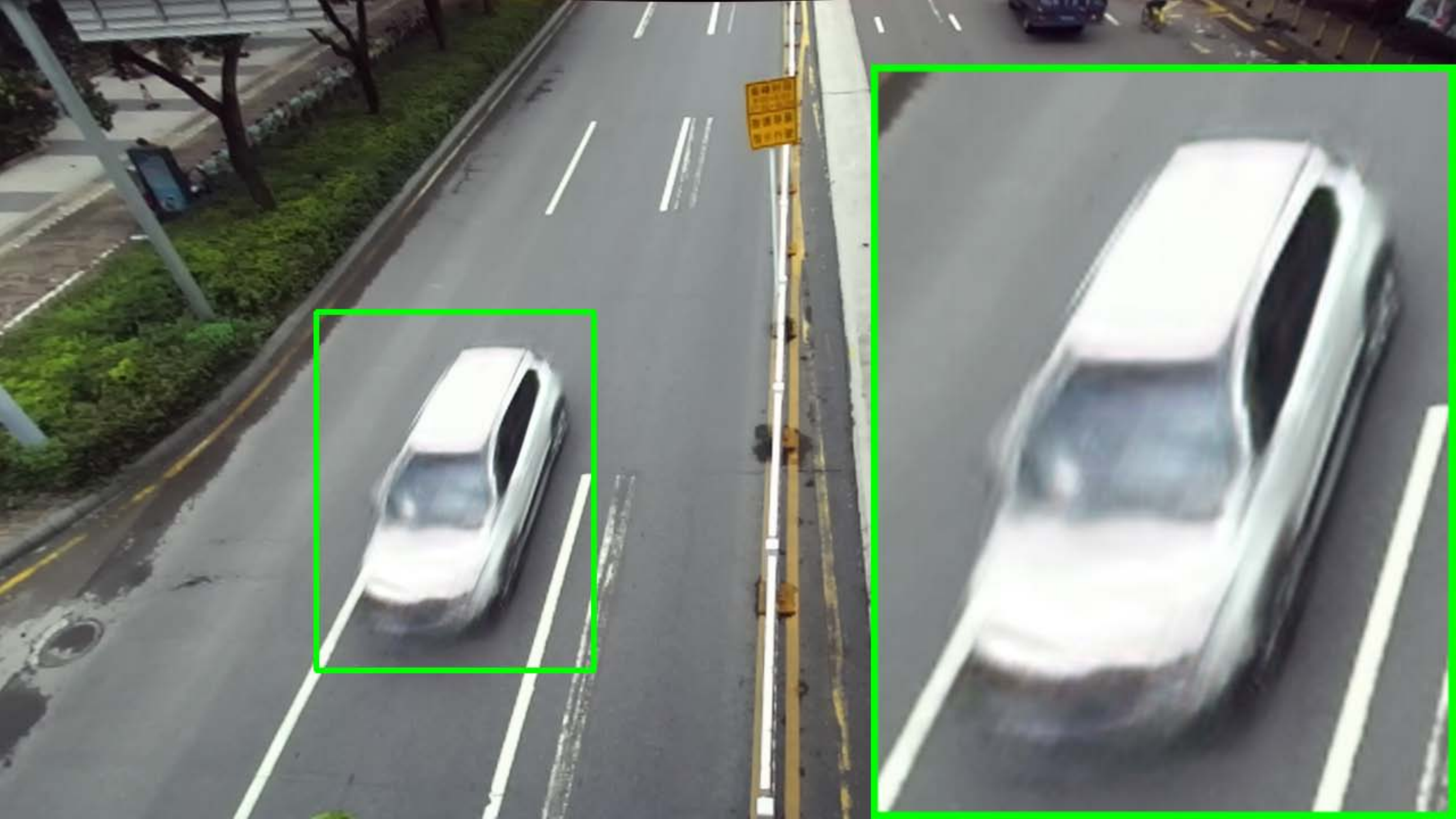} &
			\includegraphics[width=0.19\linewidth]{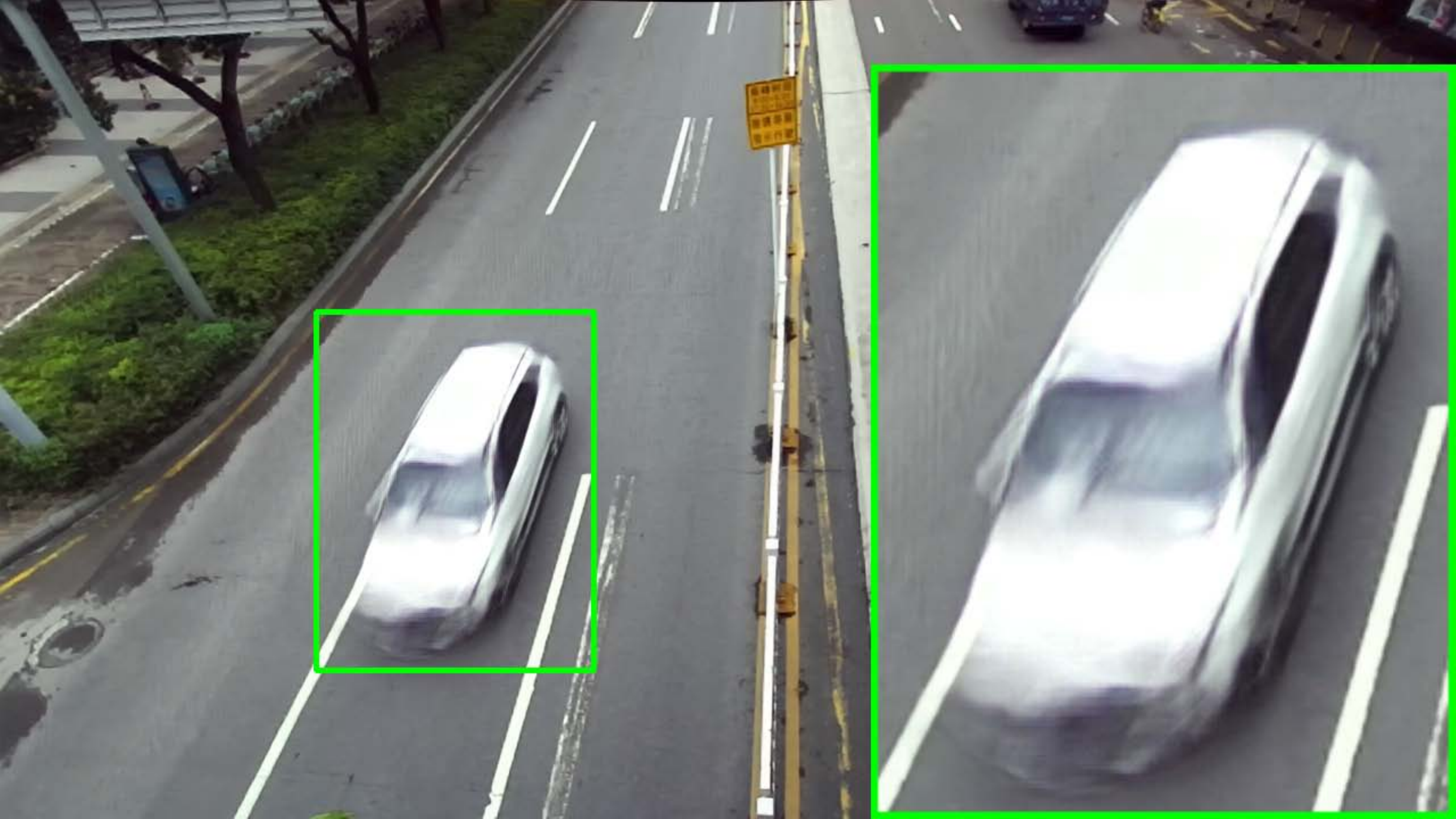}\\
			(a) Blurry image & (b) Hu \textit{et al.}~\cite{hu2014joint} & (c) Gong \textit{et al.}~\cite{gong2017motion} & (d) Nah \textit{et al.}~\cite{nah2017deep}& (e) Kupyn \textit{et al.}~\cite{kupyn2018deblurgan} \\
			PSNR / SSIM  &  21.97 / 0.8196  & 28.18 / 0.9618  &  31.54 / 0.9678 & 28.17 / 0.9394  \vspace{1.5pt}\\
			\includegraphics[width=0.19\linewidth]{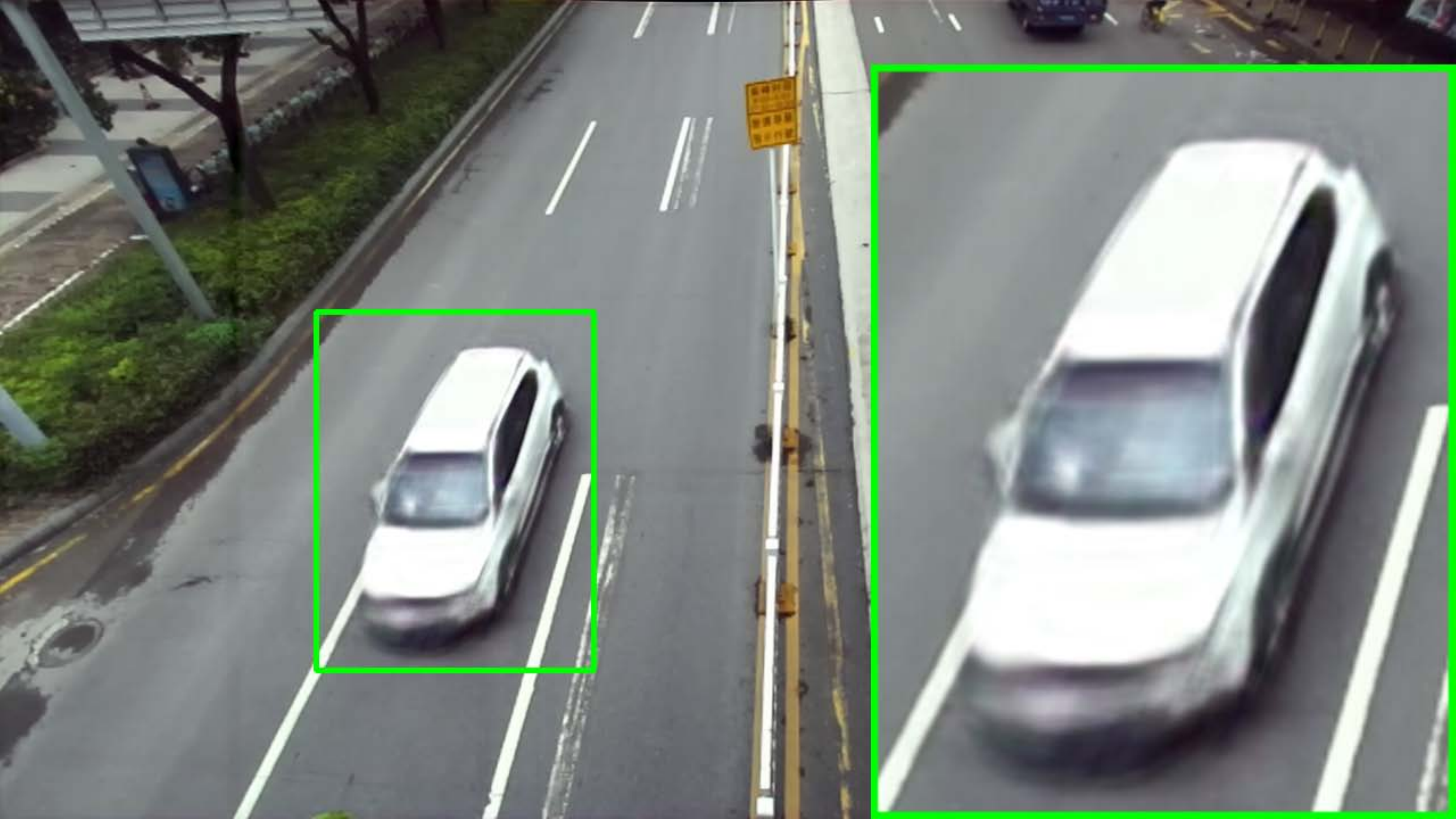} &
			\includegraphics[width=0.19\linewidth]{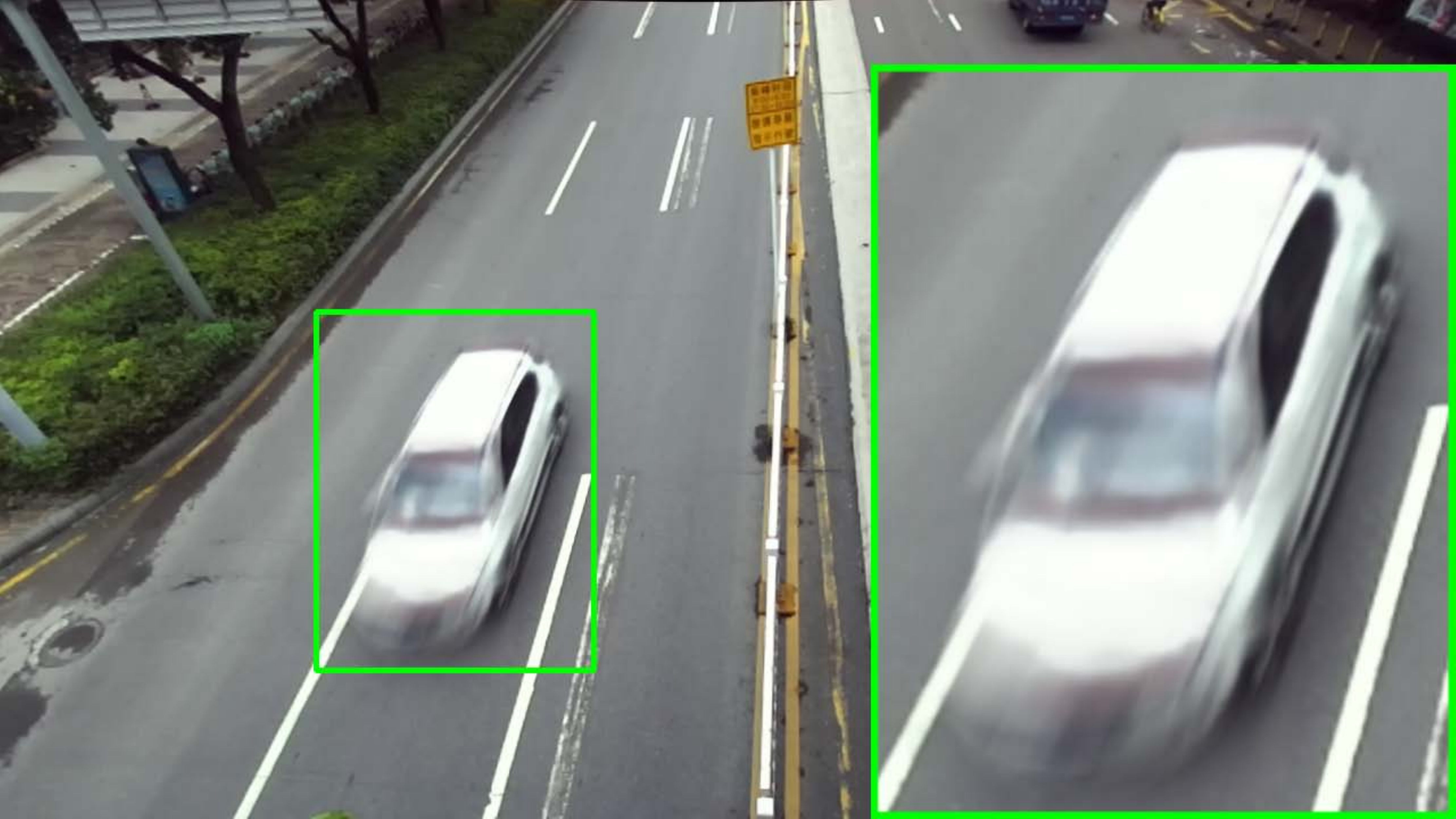} &
			\includegraphics[width=0.19\linewidth]{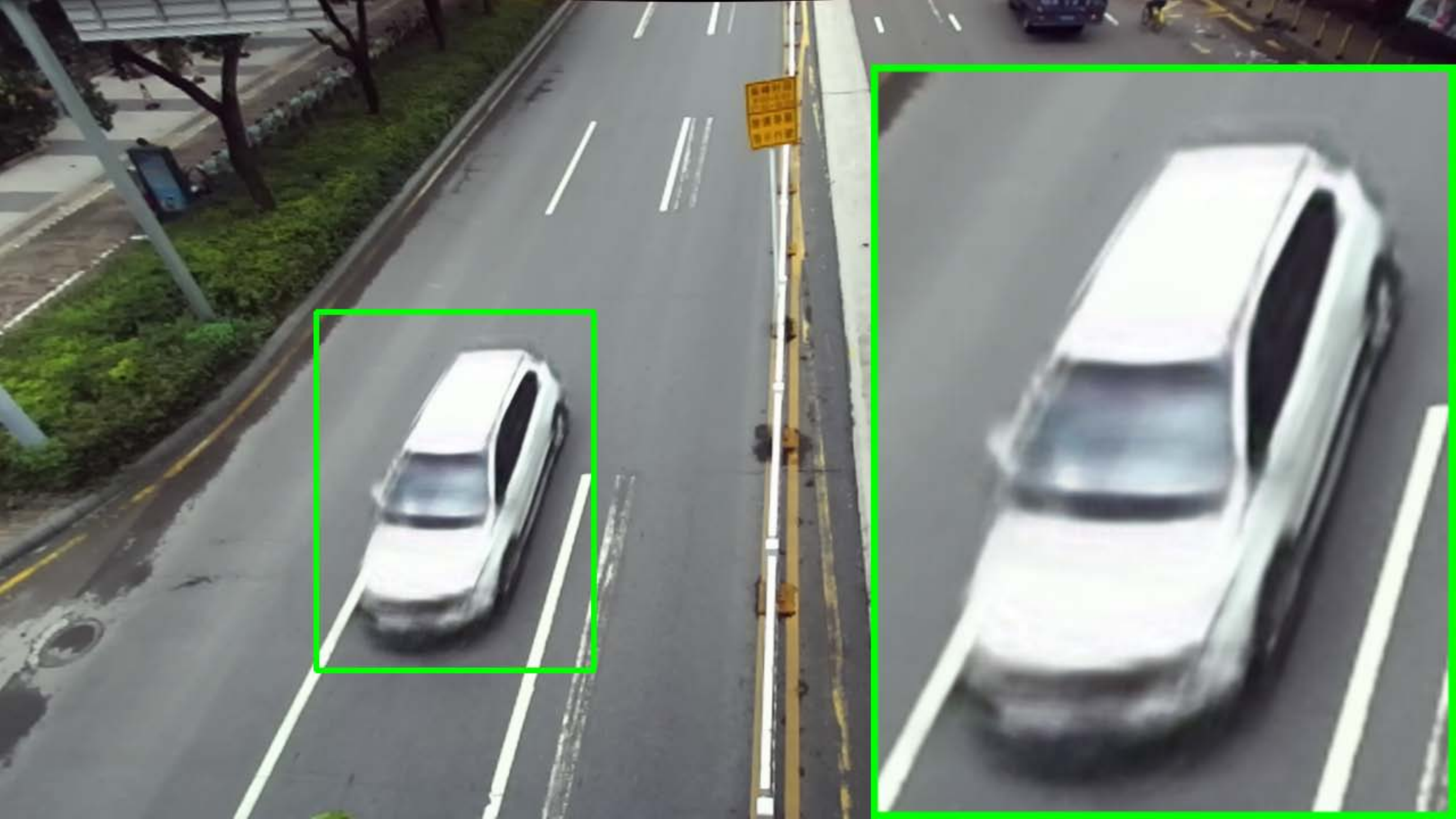} &
			\includegraphics[width=0.19\linewidth]{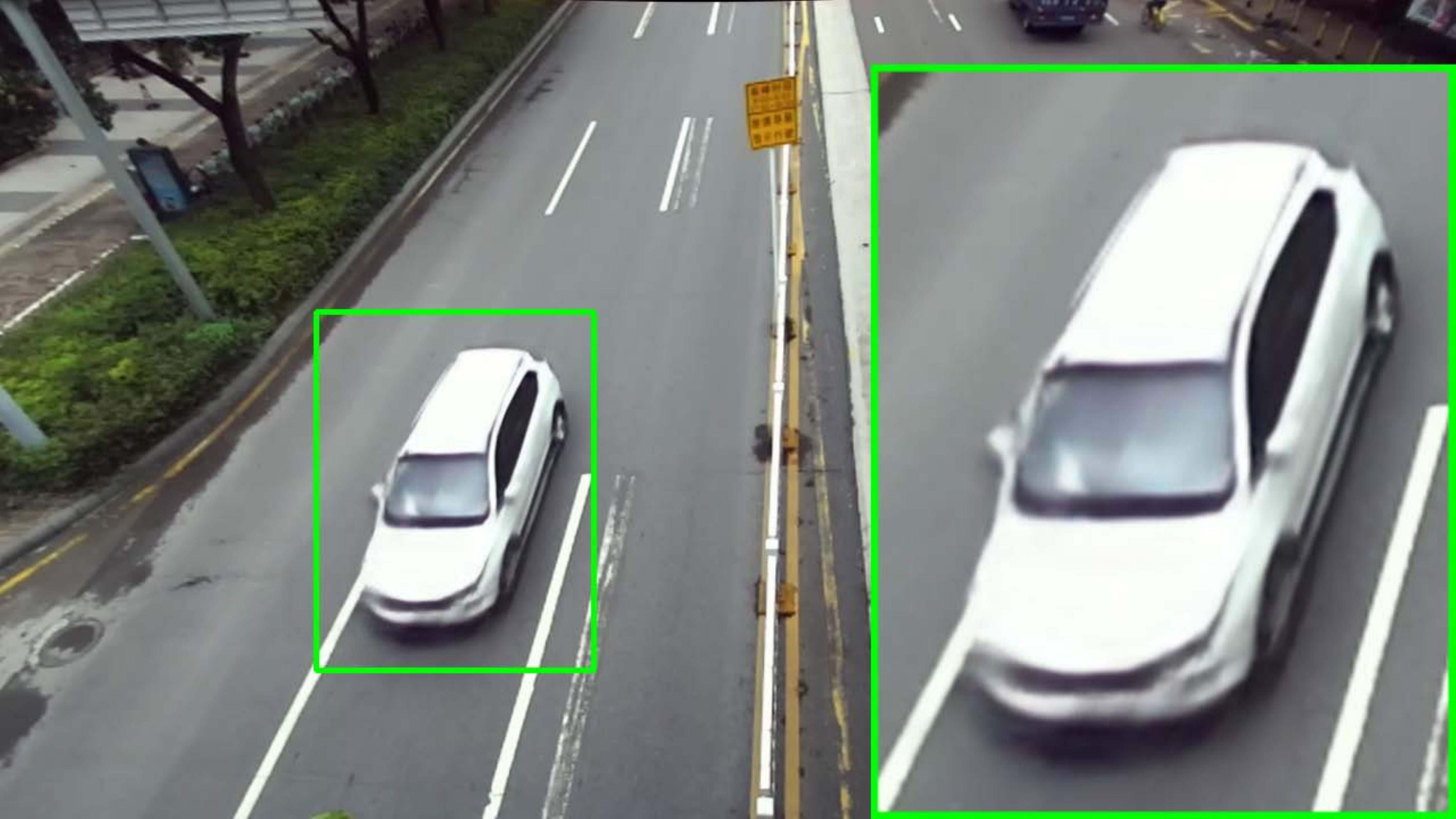} &
			\includegraphics[width=0.19\linewidth]{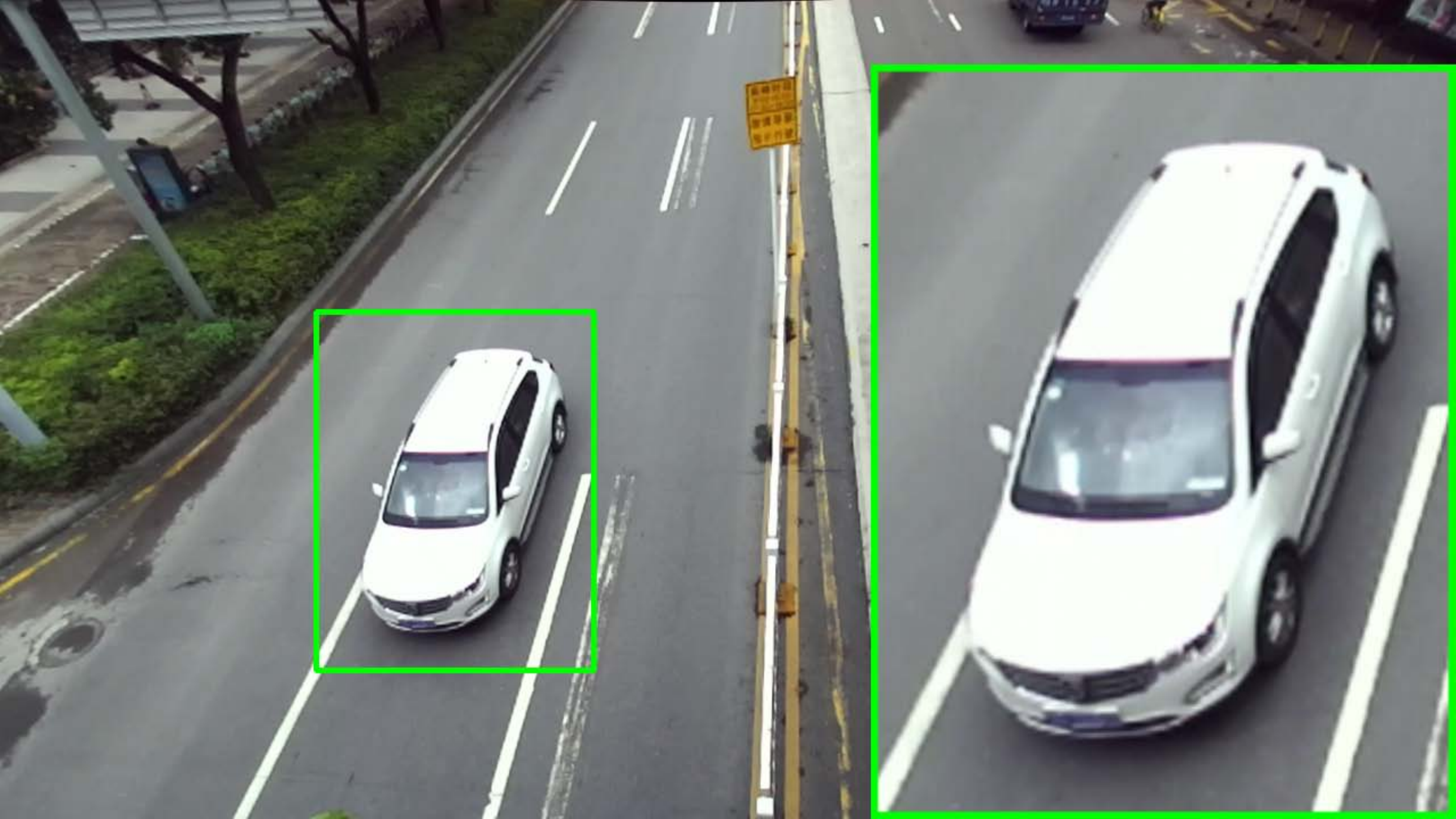}\\
			(f) Zhang \textit{et al.}~\cite{zhang2018dynamic} & (g) Tao \textit{et al.}~\cite{tao2018scale} & (h) Ours-Single & (i) Ours-Stereo & (j) Ground Truth \\
			32.61 / 0.9708  &  30.80 / 0.9732  & 31.08 / 0.9733  & \textbf{34.97 / 0.9812} & $+\infty$ / 1.0  \vspace{1.5pt}\\
			% =========================================
			\includegraphics[width=0.19\linewidth]{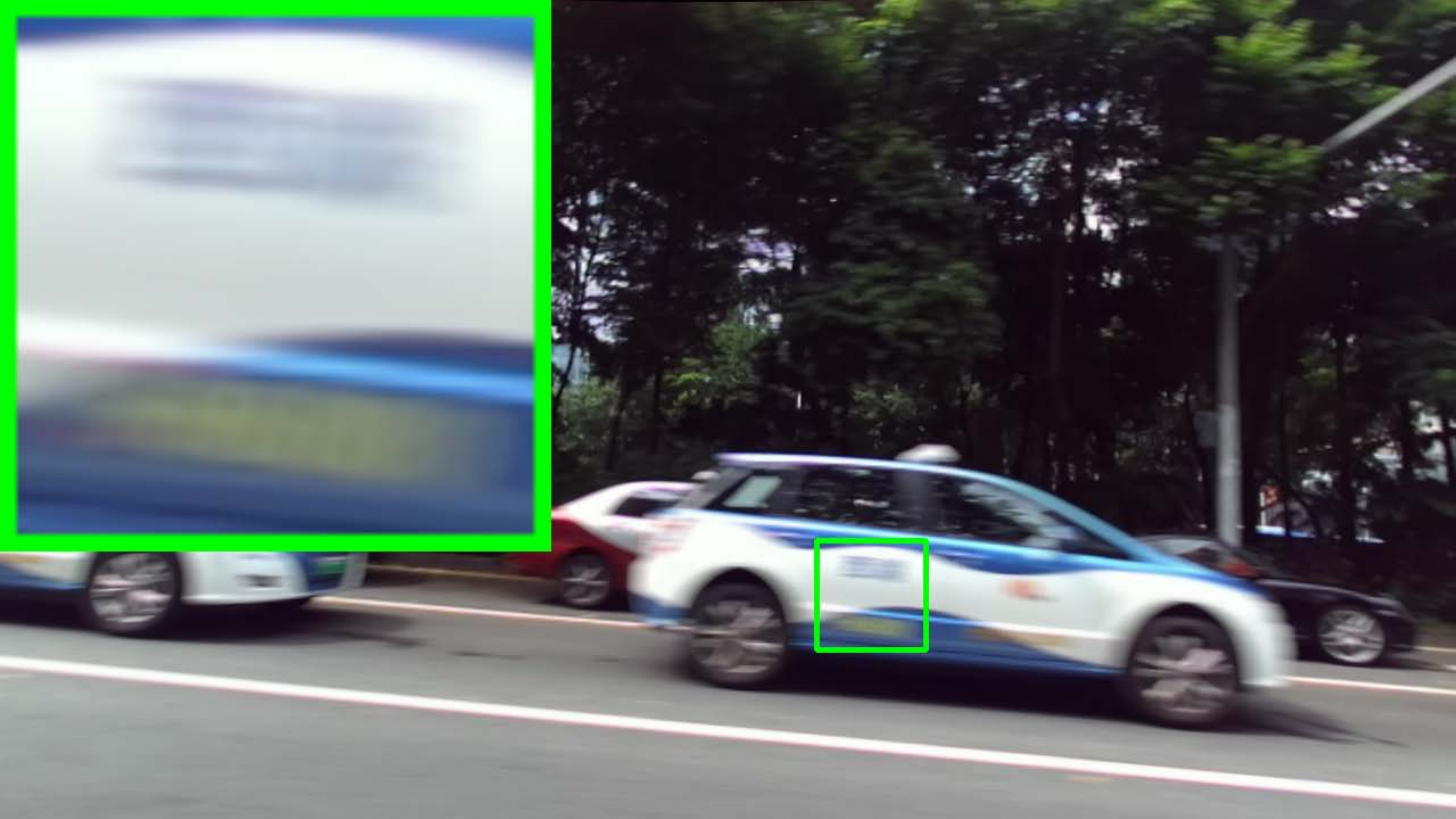} &
			\includegraphics[width=0.19\linewidth]{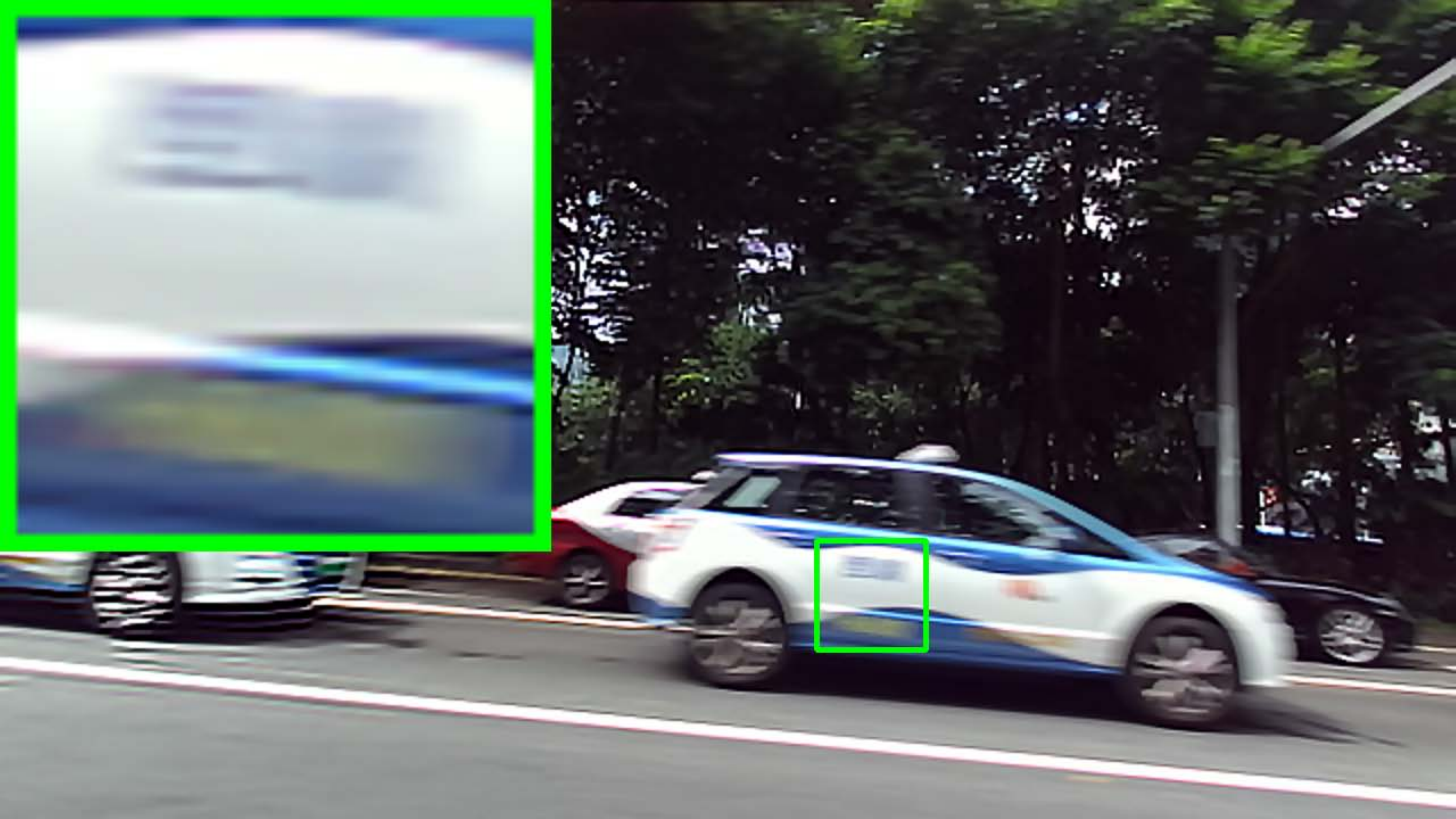} &
			\includegraphics[width=0.19\linewidth]{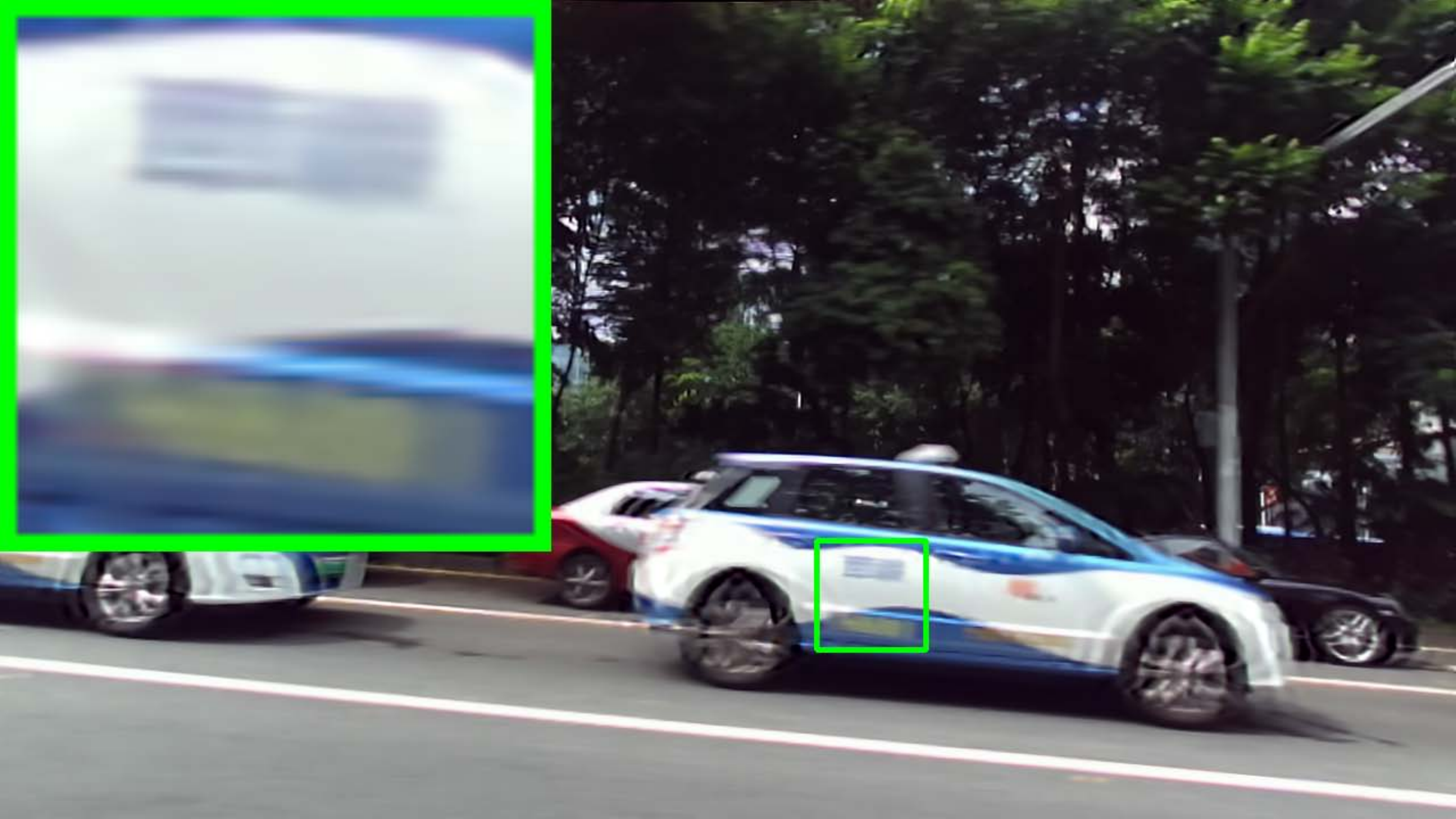} &
			\includegraphics[width=0.19\linewidth]{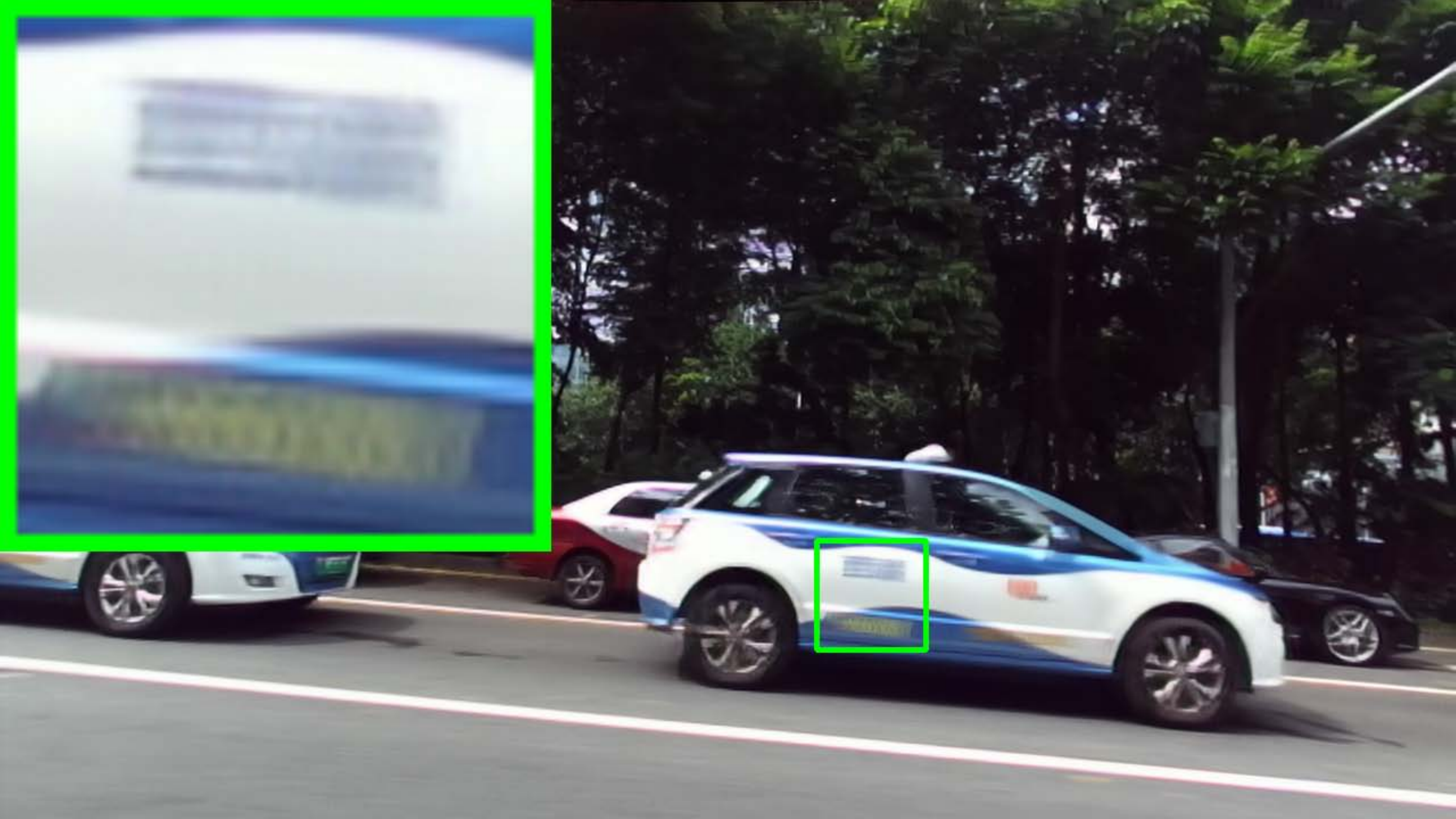} &
			\includegraphics[width=0.19\linewidth]{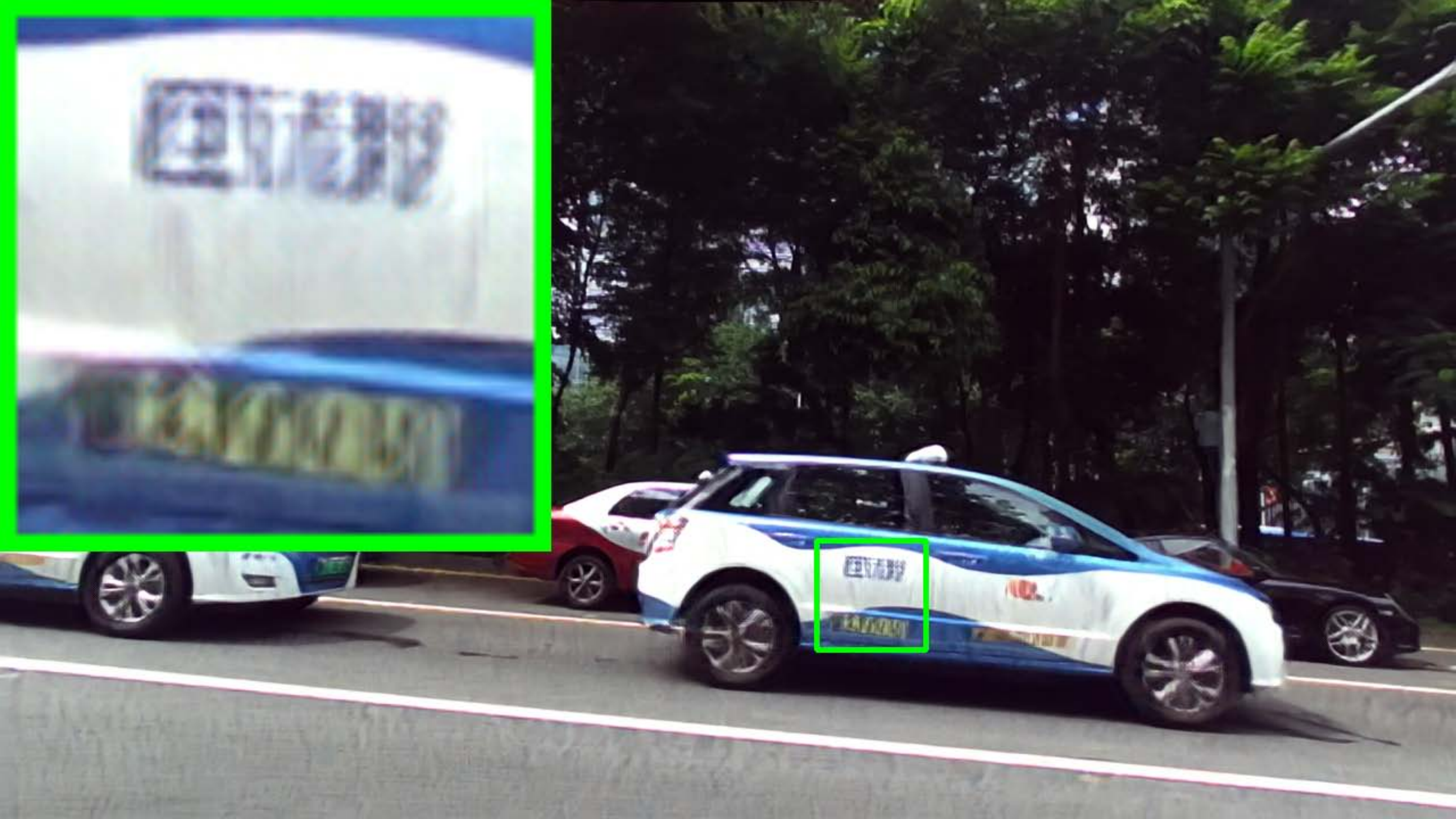}\\
			(a) Blurry image & (b) Hu \textit{et al.}~\cite{hu2014joint} & (c) Gong \textit{et al.}~\cite{gong2017motion} & (d) Nah \textit{et al.}~\cite{nah2017deep}& (e) Kupyn \textit{et al.}~\cite{kupyn2018deblurgan} \\
			PSNR / SSIM  & 20.56 / 0.7664  & 25.00 / 0.8801  &  29.76 / 0.9119 & 27.26 / 0.8619  \vspace{1.5pt}\\
			\includegraphics[width=0.19\linewidth]{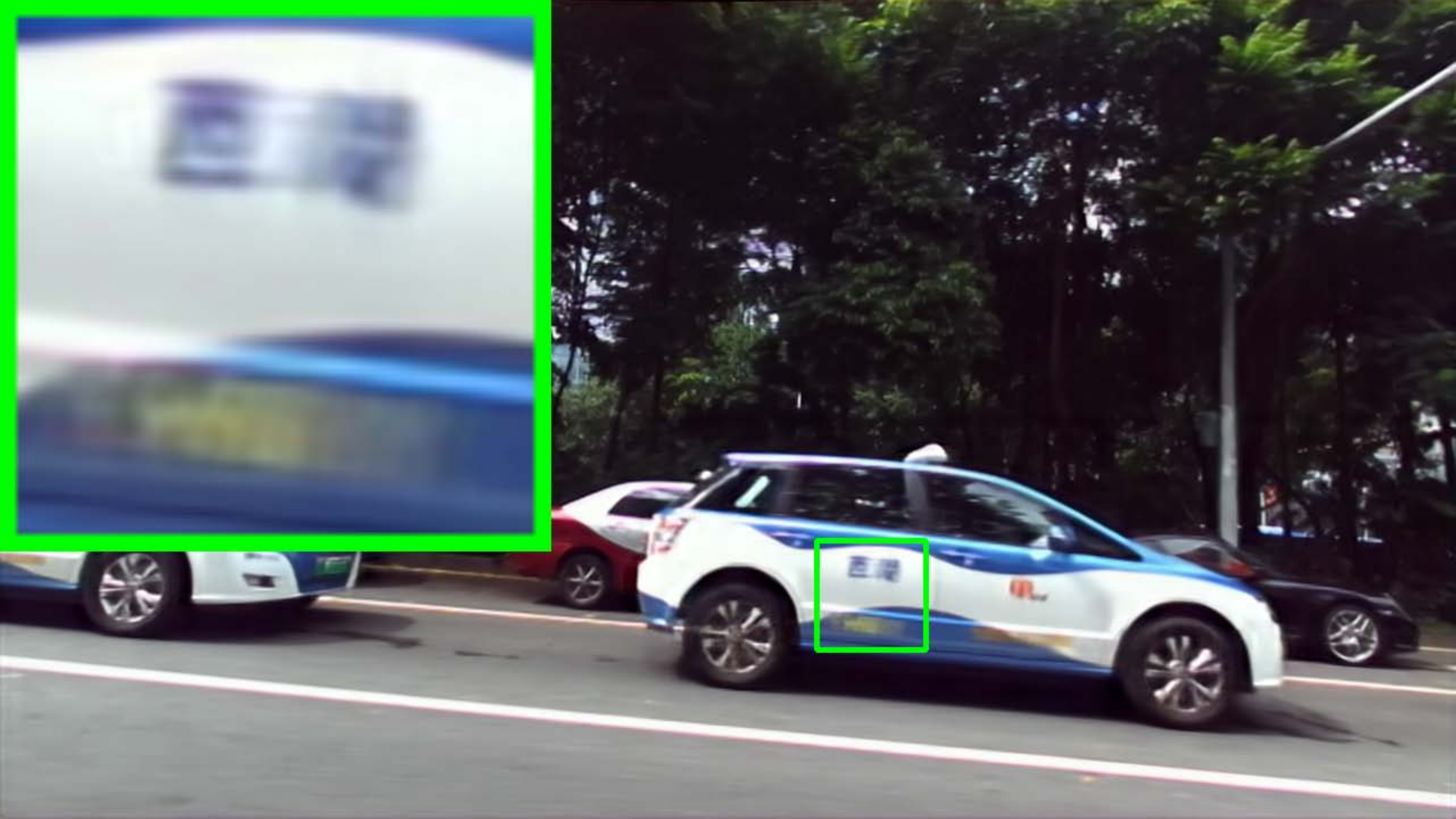} &
			\includegraphics[width=0.19\linewidth]{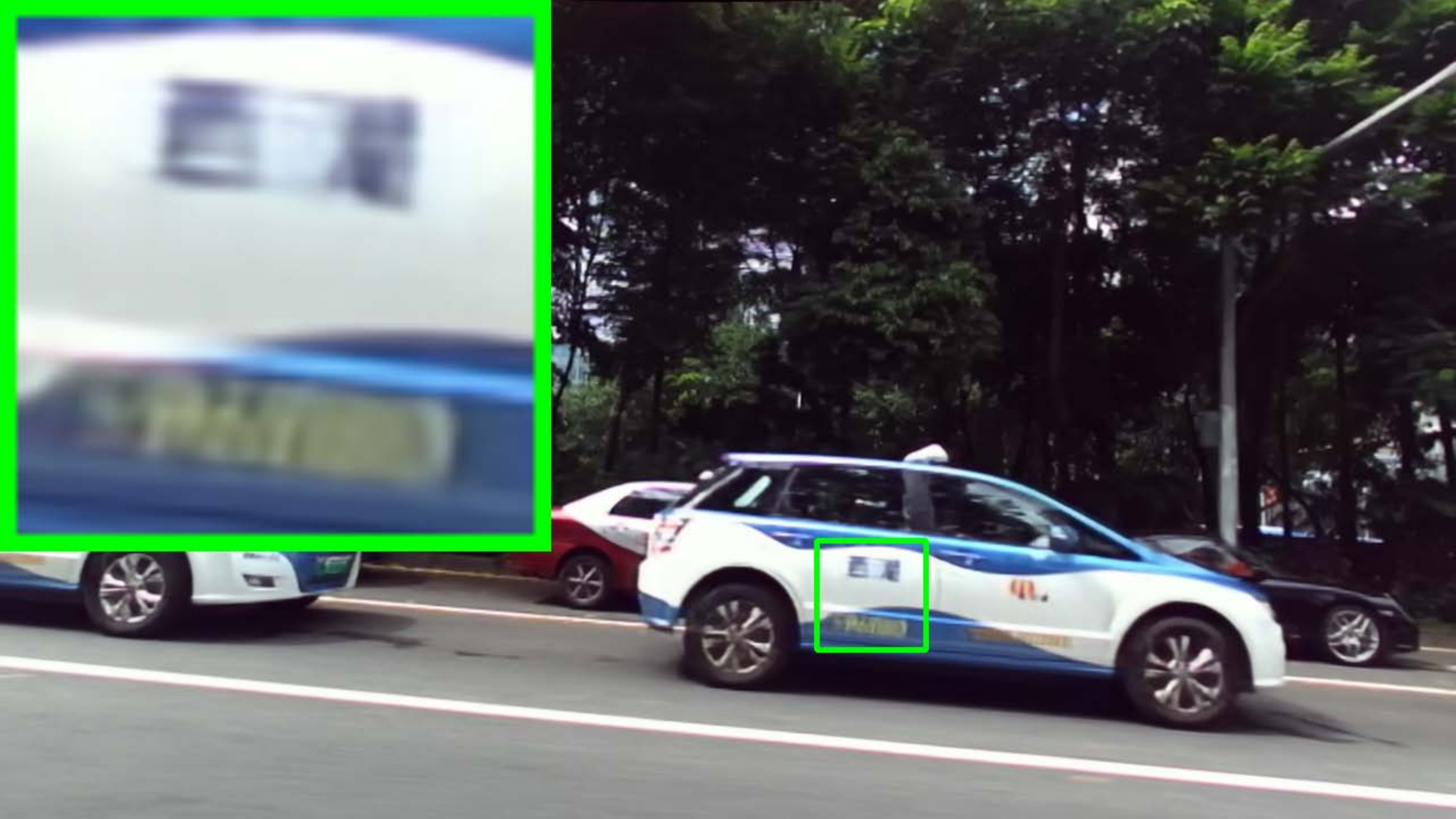} &
			\includegraphics[width=0.19\linewidth]{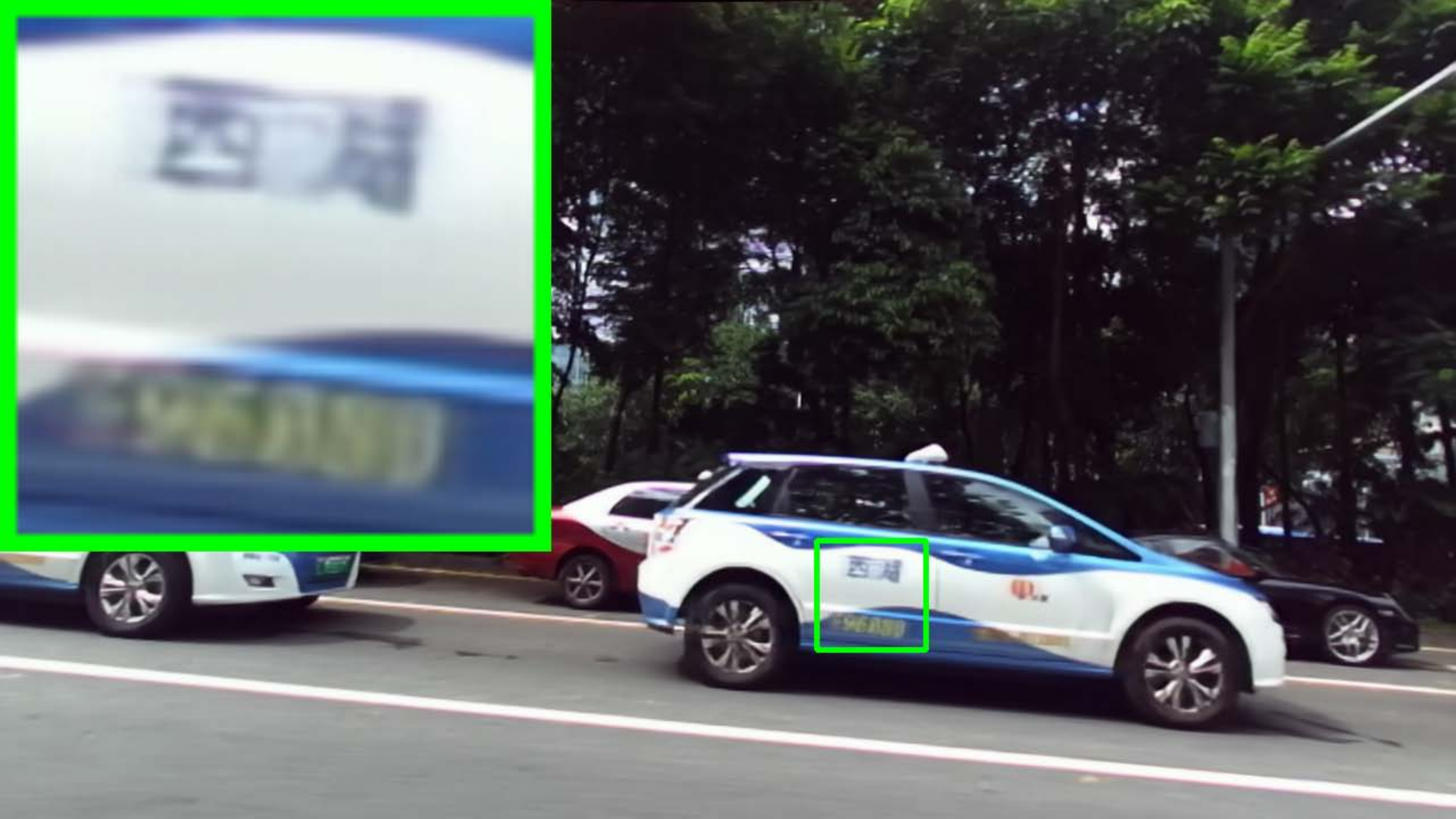} &
			\includegraphics[width=0.19\linewidth]{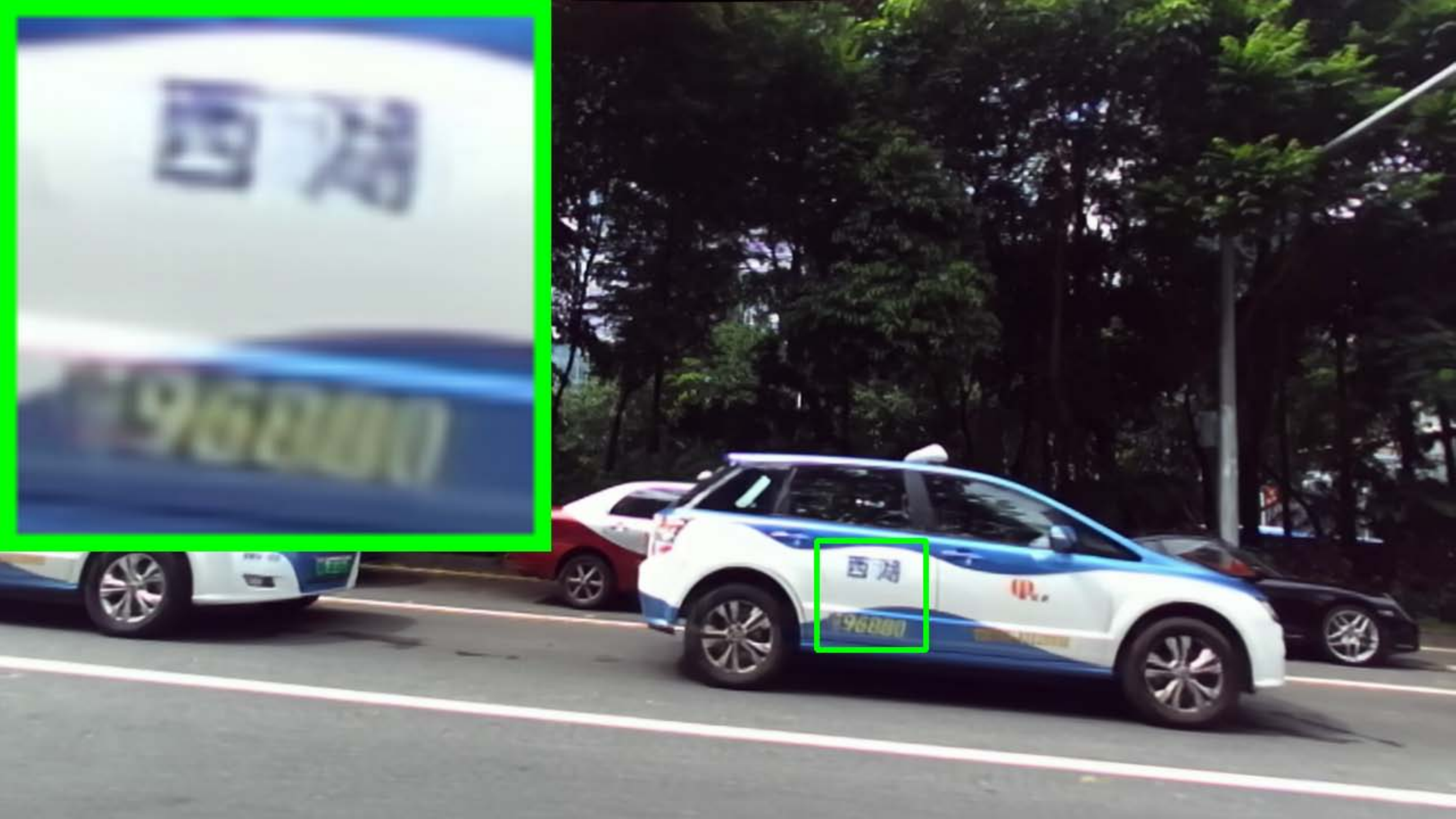} &
			\includegraphics[width=0.19\linewidth]{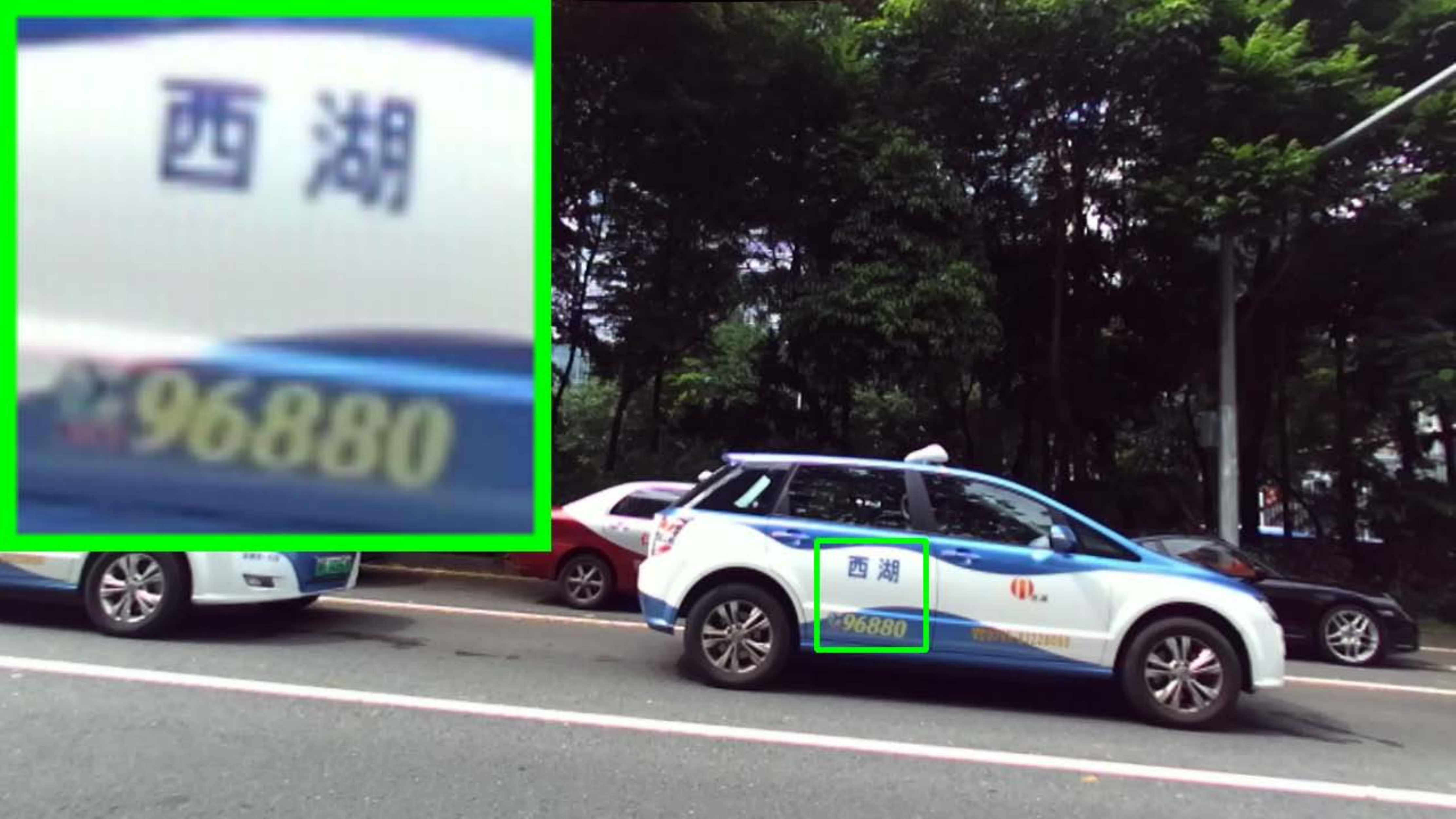}\\
			(f) Zhang \textit{et al.}~\cite{zhang2018dynamic} & (g) Tao \textit{et al.}~\cite{tao2018scale} & (h) Ours-Single & (i) Ours-Stereo & (j) Ground Truth \\
			29.82 / 0.9149  &  30.72 / 0.9284  & 31.59 / 0.9364  & \textbf{32.46 / 0.9445} & $+\infty$ / 1.0 \\
		\end{tabular}
	\end{center}
	\vspace{-2mm}
	\caption{Qualitative evaluations on our Stereo Blur Dataset. The proposed method generates much sharper images with higher PSNR and SSIM values.}
	\label{fig:show1}
	\vspace{-3mm}
\end{figure*}

% ===========
We quantitatively and qualitatively evaluate our single and stereo image deblurring networks (\textit{DeblurNet} and \textit{DAVANet}) on our dataset and compare them with the state-of-the-art deblurring algorithms, including conventional non-uniform deblurring algorithm~\cite{whyte2012non}, and CNN-based deblurring methods~\cite{sun2015learning, gong2017motion, nah2017deep, kupyn2018deblurgan, zhang2018dynamic, tao2018scale} in terms of PSNR and SSIM. 
To compare with other end-to-end CNN methods~\cite{nah2017deep, kupyn2018deblurgan, zhang2018dynamic, tao2018scale}, we fully finetune their networks on our dataset until convergence with their released codes.
For further comparison, we evaluate our single image deblurring network \textit{DeblurNet} on GOPRO dataset~\cite{nah2017deep} and compare it with aforementioned end-to-end CNN models.
% ============

%=============
\noindent \textbf{Stereo blur dataset.}
Although both \cite{nah2017deep} and \cite{tao2018scale} propose to use multi-scale recurrent scheme to improve the performance, it inevitably increases the computational cost.
To solve this problem, we apply to use two atrous residual blocks and a \textit{Context Module} to obtain the richer feature without a large network in the proposed \textit{DeblurNet}.
Table~\ref{tab:stereo_psnr_time_size} shows that \textit{DeblurNet} outperforms other state-of-the-art single-image deblurring algorithms under the proposed Stereo Blur Dataset.
Although the proposed \textit{DeblurNet} performs well with single view, we further evaluate the proposed stereo deblurring network \textit{DAVANet} with other algorithms in Table~\ref{tab:stereo_psnr_time_size}.
It demonstrates that the proposed \textit{DAVANet} performs better than the existing dynamic scene methods due to additional depth-aware and view-aggregated features.
% ============

% ============
Figure~\ref{fig:show1} shows several examples from the our testing sets.
The existing methods \cite{gong2017motion, nah2017deep, kupyn2018deblurgan, zhang2018dynamic, tao2018scale} cannot perfectly remove the large blur as depth information is not considered in their networks.
Although depth information is used in \cite{hu2014joint}, it is hard to estimate it accurately from a single image.
In this way, their estimated blur kernels are ineffective and will introduce undesired artifacts into restored images.
The proposed \textit{DAVANet} estimates disparity considered as non-uniform prior information to handle spatially variant blur in dynamic scenes.
Moreover, it also fuses two-view varying information, which provides more effective and additional information for deblurring.
With depth awareness and view aggregation, Figure~\ref{fig:show1} shows our proposed \textit{DAVANet} can restore sharp and artifact-free images.
% ============

% ============
\begin{figure*}[t]\footnotesize
	\centering
	\renewcommand{\tabcolsep}{1.2pt}
	\renewcommand{\arraystretch}{1}
	\begin{center}
		\begin{tabular}{cccccccc}
			\includegraphics[width=0.118\linewidth]{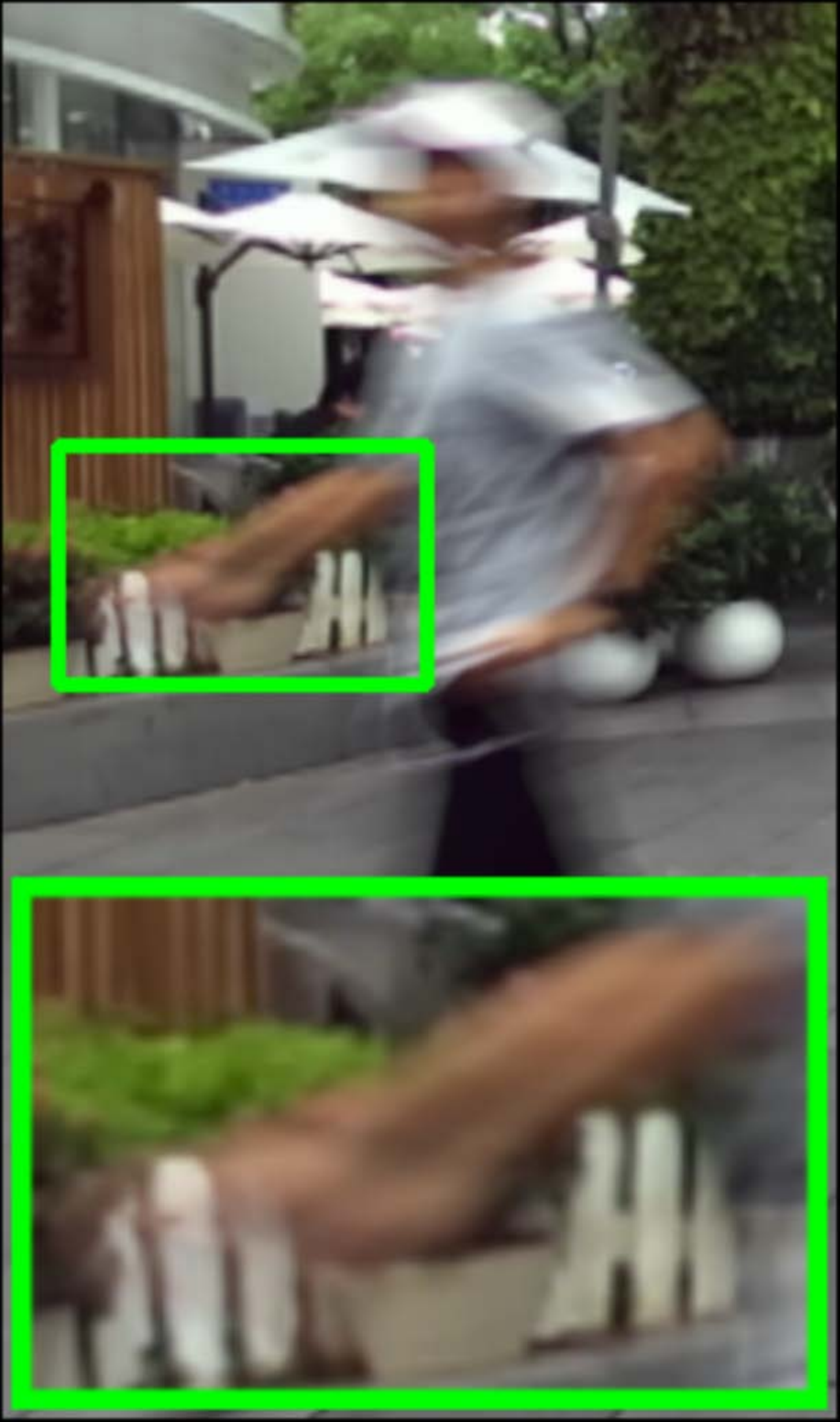} &
			\includegraphics[width=0.118\linewidth]{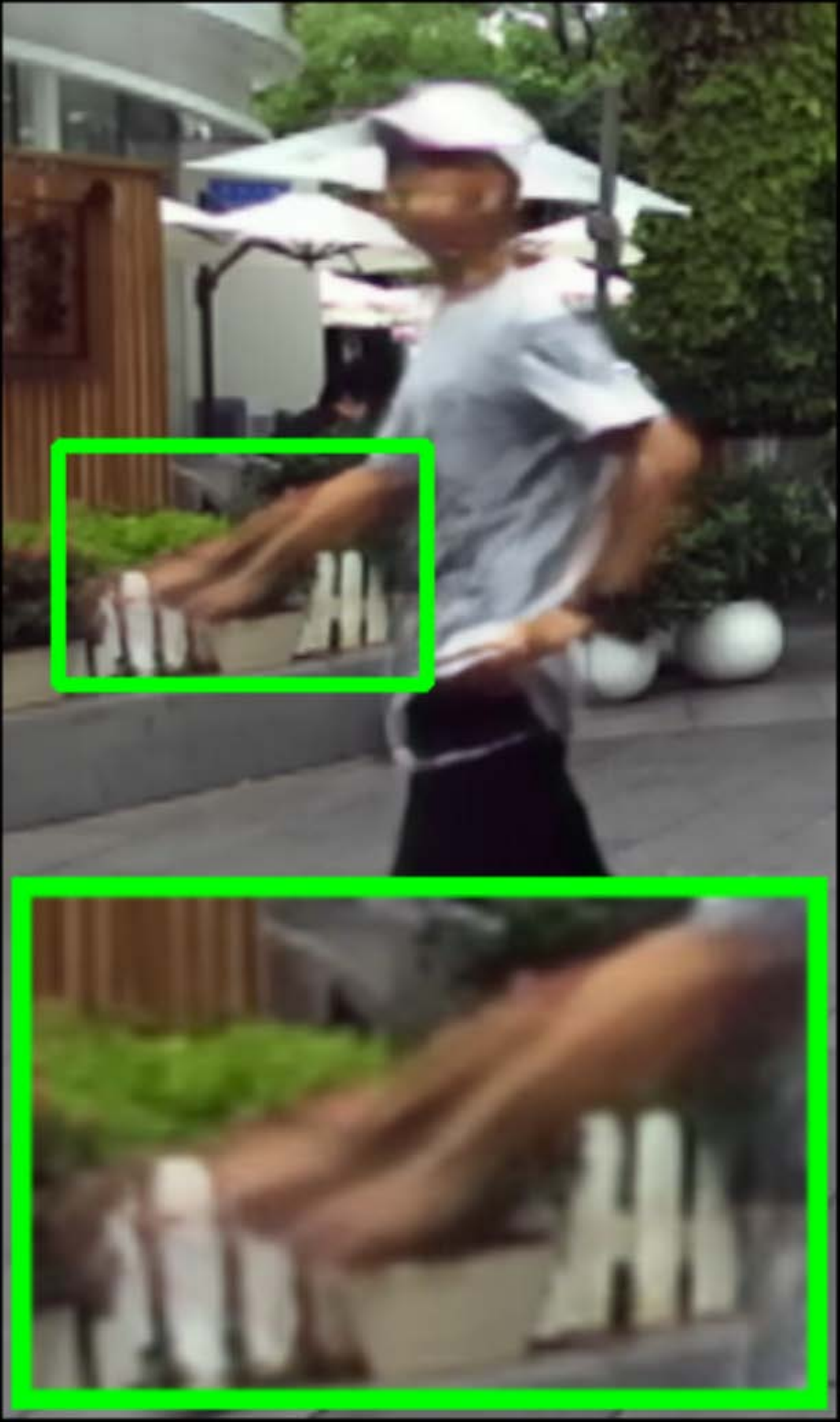} &
			\includegraphics[width=0.118\linewidth]{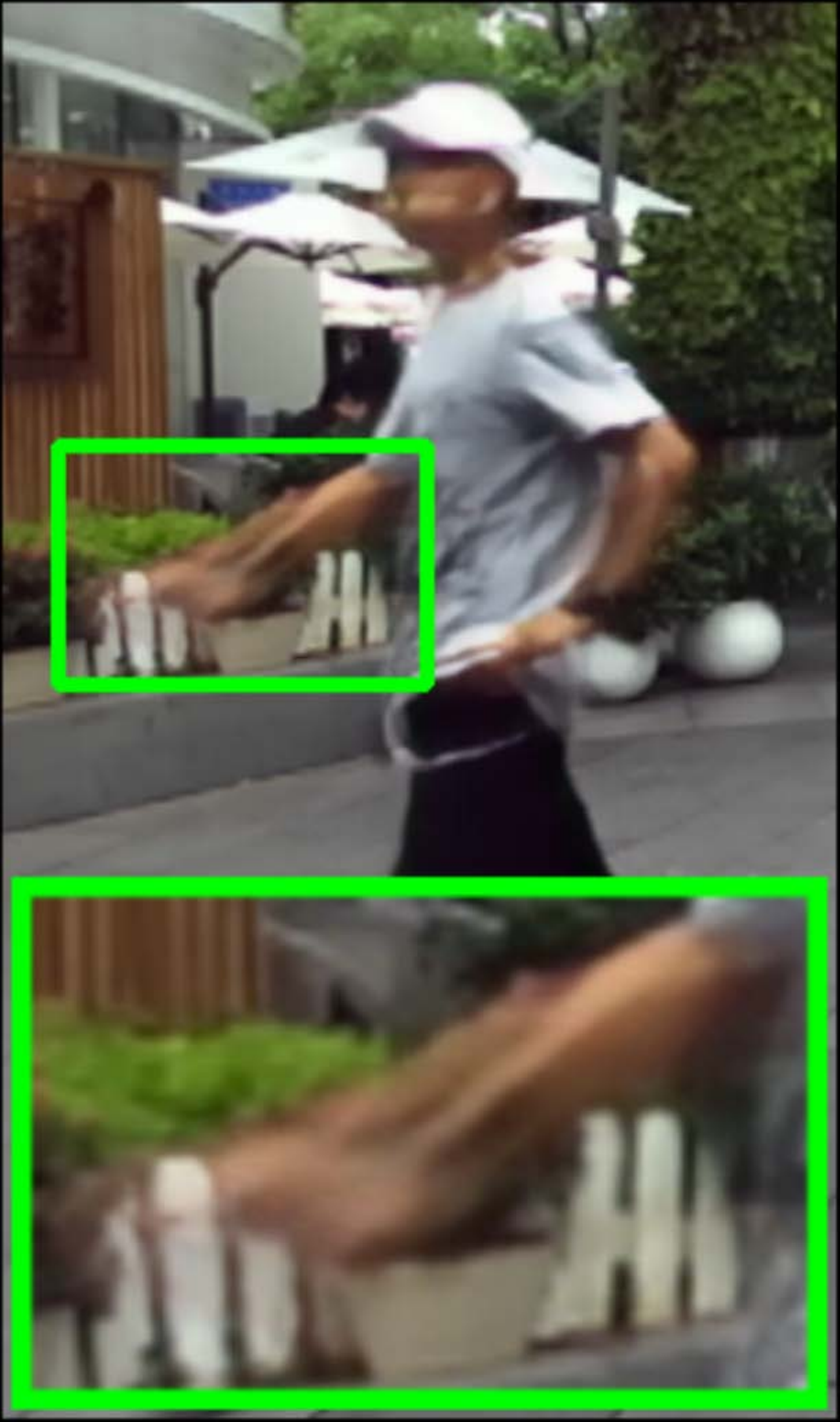} &
			\includegraphics[width=0.118\linewidth]{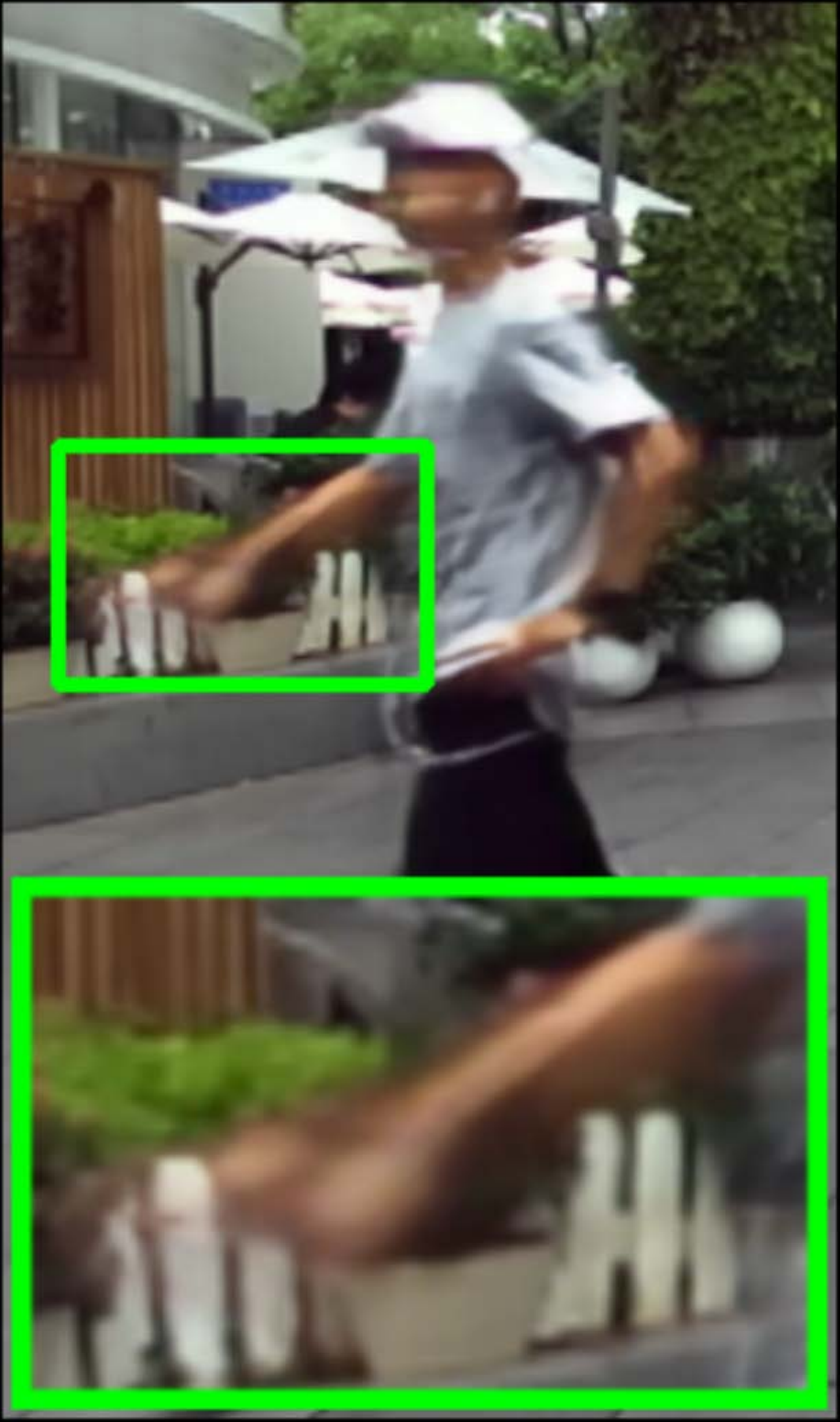} &
			\includegraphics[width=0.118\linewidth]{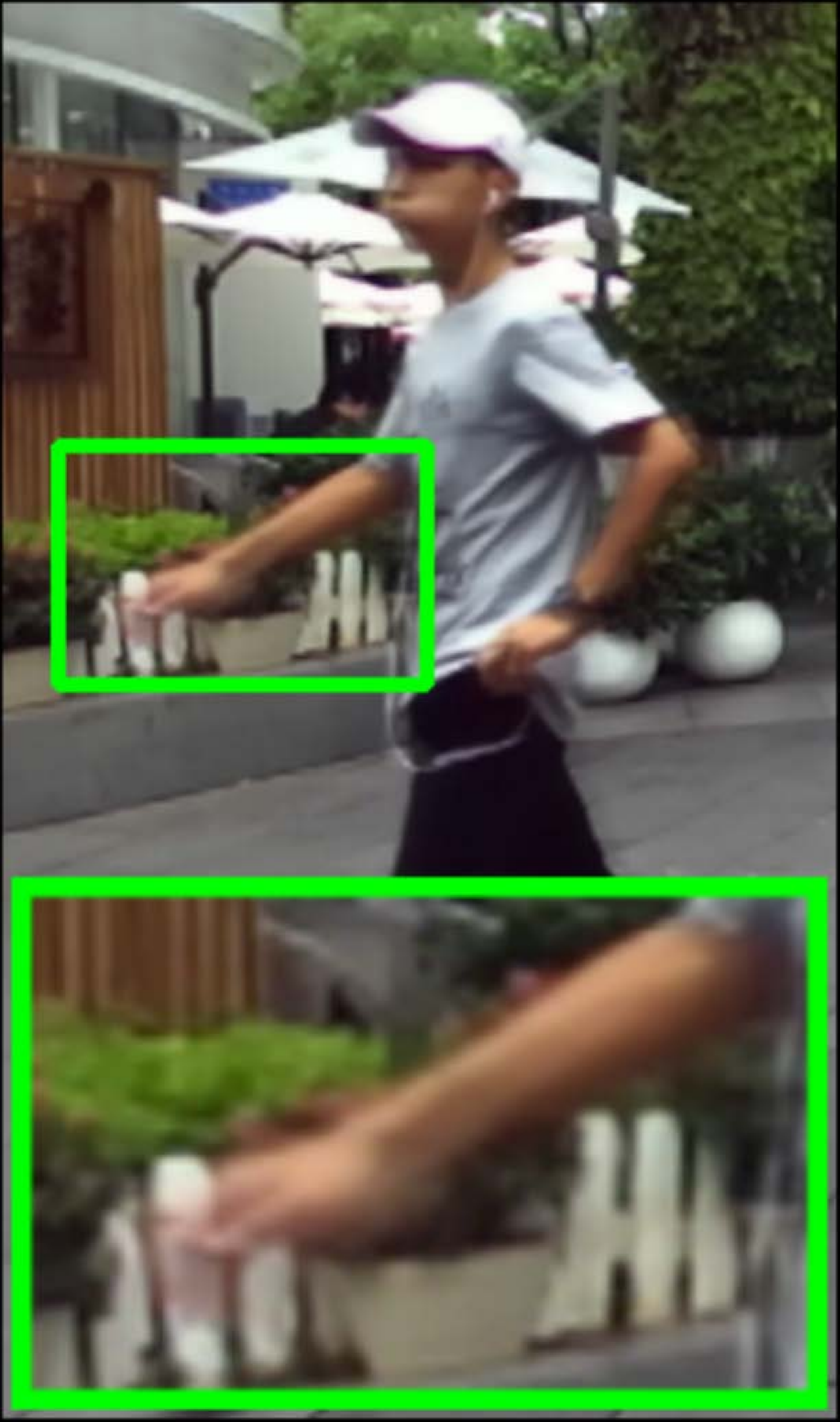}&
			\includegraphics[width=0.118\linewidth]{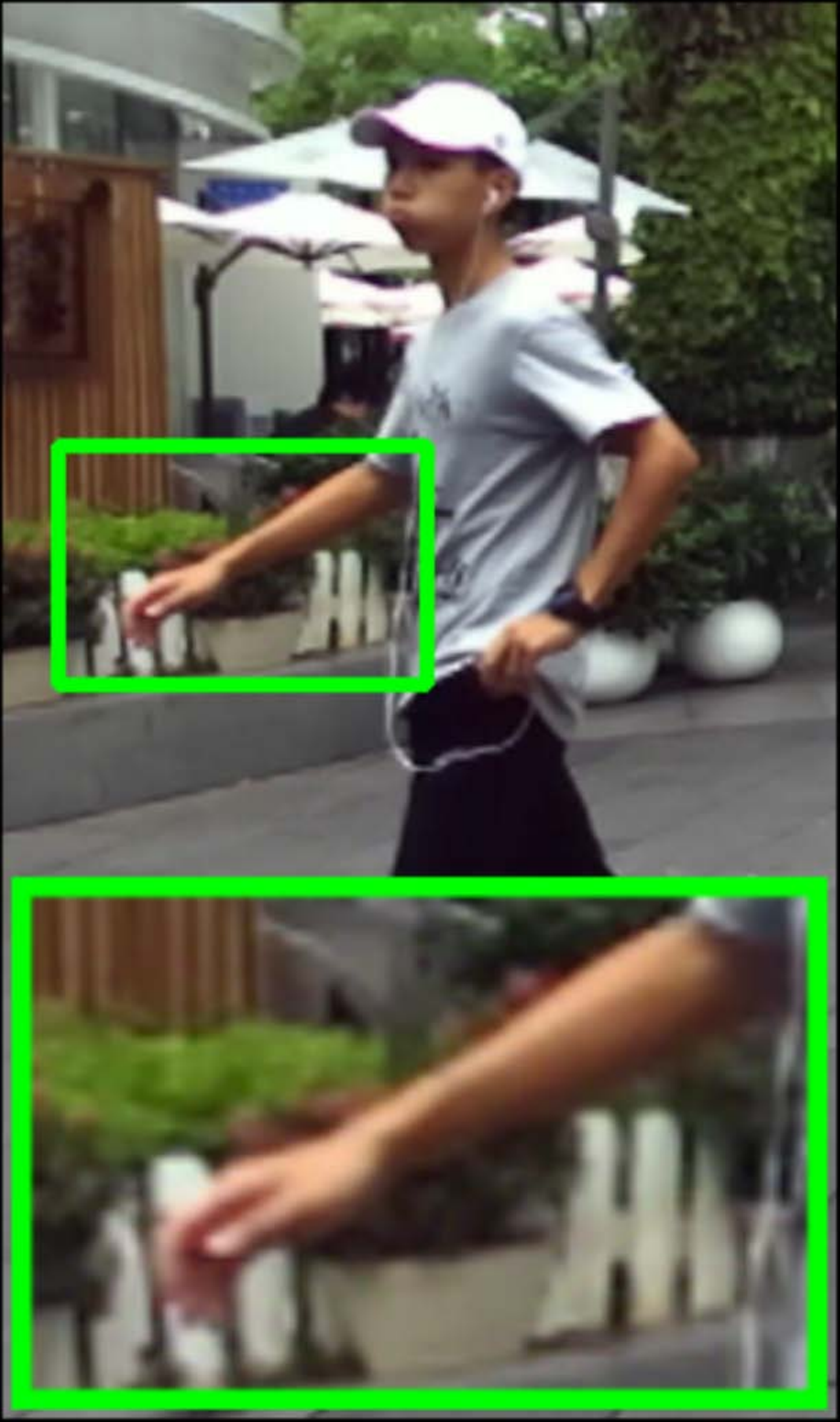} &
			\includegraphics[width=0.118\linewidth]{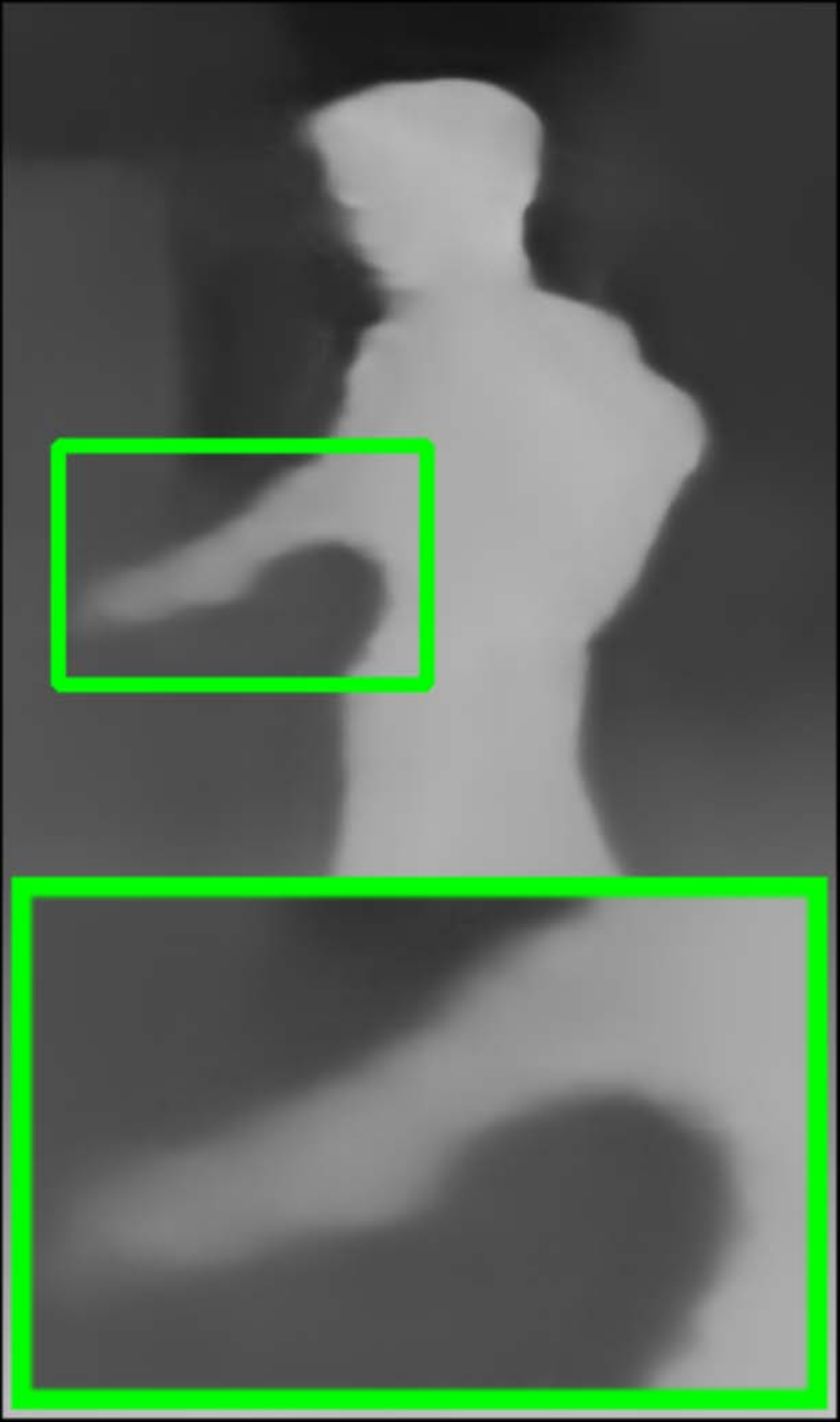}&
			\includegraphics[width=0.118\linewidth]{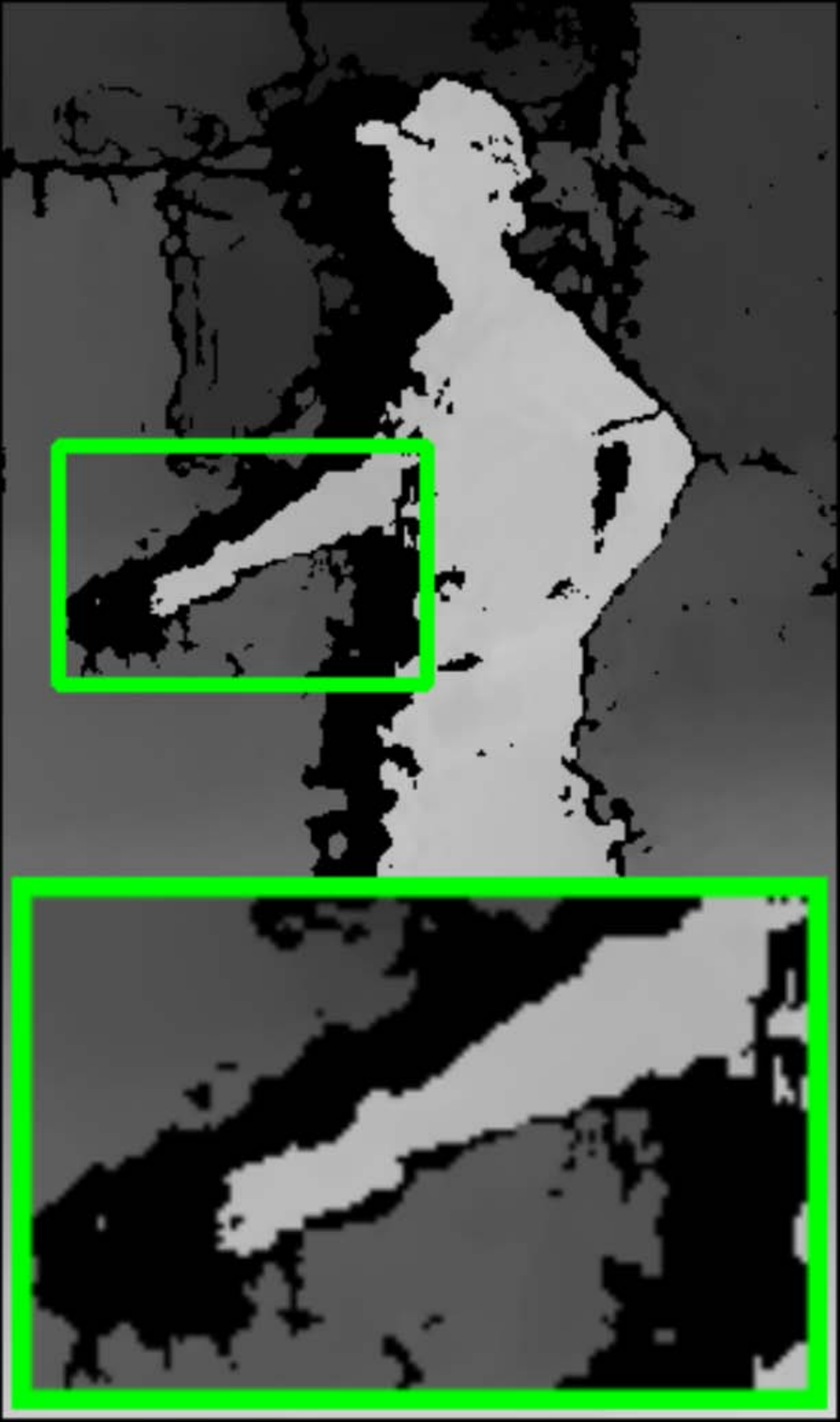}\\
			(a) Blurry image & (b) Single  & (c) w/o \textit{DAVA} & (d) w/o \textit{VA} & (e) Stereo (\textit{DAVA}) & (f) Ground Truth & (g) Disparity (P)& (h) Disparity (T)\\
		\end{tabular}
	\end{center}
	\vspace{-2mm}
	\caption{
		The effectiveness of disparity.
		(a), (f), (g) and (h) denote the blurry image, sharp image, our predicted disparity and ground truth disparity, respectively.
		(b) and (e) are the deblurring results from the proposed single image deblurring network \textit{DeblurNet} and stereo deblurring network \textit{DAVANet}.
		In (c), taking two left images as input, \textit{DispBiNet} cannot provide any depth information or disparity for depth awareness and view aggregation.
		In (d), to only remove the effect of view aggregation, we do not warp the feature from the other view in the \textit{FusionNet}.
		As the proposed network can estimate accurate disparities and make use of them, it outperforms to the other configurations.
	}
	\label{fig:depth-analysis}
	\vspace{-1.18mm}
\end{figure*}
% ============

% ============
\noindent \textbf{GOPRO dataset.}
Though our single image deblurring network \textit{DeblurNet} performs well on our dataset, we further evaluate it on public GOPRO dataset~\cite{nah2017deep} and compare it with the state-of-the-art CNN models.
According to Table~\ref{tab:psnr_gopro}, the proposed \textit{DeblurNet} with small size outperforms other algorithms in terms of PSNR and SSIM,
which further demonstrates the effectiveness of \textit{Context Module}.
% ============
\begin{table}[h]
	\centering
	\caption{Quantitative evaluation on the GOPRO dataset \cite{nah2017deep}, in terms of PSNR and SSIM. }
	\resizebox{\linewidth}{!} {
		\begin{tabular}{lcccccc}
			\toprule
			Method   
			& Nah~\cite{nah2017deep}
			& Kupyn~\cite{kupyn2018deblurgan}
			& Zhang~\cite{zhang2018dynamic}
			& Tao~\cite{tao2018scale}
			& Ours-Single\\
			\midrule
			PSNR     & 28.49      & 25.86      & 29.19      & 30.26      & \bf{30.55}  \\
			SSIM      & 0.9165    & 0.8359    & 0.9306    & 0.9342    & \bf{0.9400}  \\
			\bottomrule
		\end{tabular}
	}
	\label{tab:psnr_gopro}
	\vspace{-3mm}
\end{table}
% ============

% ============
\noindent \textbf{Running time and model size.}
We implement our network using PyTorch platform~\cite{pytorch}. 
To compare running time, we evaluate the proposed method and state-of-the-art image deblurring methods on the same server with an Intel Xeon E5 CPU and an NVIDIA Titan Xp GPU.
As traditional blind or non-blind algorithms are used in \cite{whyte2012non, sun2015learning, gong2017motion}, their methods are time-consuming.
With GPU implementation, deep learning-based methods \cite{nah2017deep, kupyn2018deblurgan, zhang2018dynamic, tao2018scale} are efficient.
To enlarge the receptive field, multi-scale recurrent scheme and large CNN kernel size (e.g. $5\times5$) are used in \cite{nah2017deep, tao2018scale}.
For the same purpose, spatially variant RNNs are used in \cite{zhang2018dynamic}.
They all lead to long computation time.
We find that the proposed \textit{Context Module}, which utilizes convolutions with different dilation rates, can embed multi-scale features and enlarge the receptive field at a low computational cost.
In addition, only $3\times3$ convolutional layers are used in the proposed network which further reduces the size of network.
According to Table~\ref{tab:stereo_psnr_time_size}, the proposed network is more efficiency with a small model, compared to the existing CNN-based methods.
% ============
\subsection{Analysis and Discussions}
\noindent \textbf{Effectiveness of the disparity.}
The proposed model \textit{DAVANet} utilizes estimated disparities in two ways: Depth Awareness (\textit{DA}) and View Aggregation (\textit{VA}).
To remove the effect of view aggregation, we do not warp features from the other view in the \textit{FusionNet}, as shown in Figure~\ref{fig:depth-analysis}(d).
Furthermore, to remove the effect of both depth awareness and view aggregation, we feed two exactly the same images into the proposed network, where no depth information or disparity can be obtained, as shown in Figure~\ref{fig:depth-analysis}(c).
And we also compare the proposed \textit{DAVANet} with the proposed single image network \textit{DeblurNet}, as shown in Figure~\ref{fig:depth-analysis}(b). 
The Figure~\ref{fig:depth-analysis} demonstrates that the proposed \textit{DAVANet} with depth awareness and view aggregation performs better, using the accurate disparities provided by \textit{DispBiNet}.
% ============

% ============
\noindent \textbf{Ablation study.}
The performance improvement of our proposed network should be attributed to three key components, including: \textit{Context Module}, depth awareness, and view aggregation. To demonstrate the effectiveness of each component in the proposed networks, we evaluate the following three variant networks for controlled comparison: 
(a) To validate the effectiveness of the \textit{Context Module}, we replace the \textit{Context Module} of \textit{DeblurNet} by the one-path convolution block with the same number of layers;
(b) To remove the effect of depth information, we remove disparity loss of \textit{DispBiNet} but keep the original input features to \textit{DeblurNet}, where no depth information is involved. 
The whole network is updated by deblurring losses;
(c) To remove the effect of view aggregation, we substitute the concatenation component, the view aggregated features $F_{views}^L$, with a copy of the reference view features $F^L$  in \textit{FusionNet} (refer to Figure~\ref{fig:fusion} for clarification). 
We train these networks using the same strategy as aforementioned in Section~\ref{sec:training}. Table~\ref{tab:ablation} shows the proposed network is the best when all components are adopted.
% ============

\begin{table}[h]
	\centering
	\caption{Ablation study for the effectiveness of context module, depth awareness and view aggregation. Please see text for details.}
	\resizebox{\linewidth}{!} {
		\begin{tabular}{l | cc | ccc}
			\toprule
			Network   
			& w/o Context
			& \textbf{Single}
			& w/o \textit{DA}
			& w/o \textit{VA}
			& \textbf{Stereo}\\
			\midrule
			PSNR     & 31.40  & \bf{31.97}    & 32.69     & 32.53    &\bf{33.19}\\
			SSIM      & 0.9461       & \bf{0.9507}        & 0.9569         & 0.9558    & \bf{0.9586}\\ 
			\bottomrule
		\end{tabular}
	}
	\label{tab:ablation}
	\vspace{-3mm}
\end{table}

% ============
\section{Conclusions}

In this paper, we present an efficient and effective end-to-end network, \textit{DAVANet}, for stereo image deblurring. 
The proposed \textit{DAVANet} benefits from depth awareness and view aggregation,
where the depth and two-view information are effectively leveraged for spatially-varying blur removal in dynamic scenes. 
We also construct a large-scale, multi-scene and depth-varying dataset for stereo image deblurring, which consists of 20,637 blurry-sharp stereo image pairs from 135 diverse sequences. 
The experimental results show that our network outperforms the state-of-the-art methods in terms of accuracy, speed, and model size.
\section{Acknowledgements}
This work have been supported in part by the National Natural Science Foundation of China (No. 61671182 and 61872421) and Natural Science Foundation of Jiangsu Province (No. BK20180471).

\clearpage
{\small
	\bibliographystyle{ieee}
	\bibliography{references}

\begin{thebibliography}{10}\itemsep=-1pt

\bibitem{stereolabs}
Stereolabs.
\newblock \url{https://www.stereolabs.com/}.

\bibitem{aittala2018burst}
Miika Aittala and Fr{\'e}do Durand.
\newblock Burst image deblurring using permutation invariant convolutional
  neural networks.
\newblock In {\em ECCV}, 2018.

\bibitem{chang2018pyramid}
Jia-Ren Chang and Yong-Sheng Chen.
\newblock Pyramid stereo matching network.
\newblock In {\em CVPR}, 2018.

\bibitem{chen2018stereoscopic}
Dongdong Chen, Lu Yuan, Jing Liao, Nenghai Yu, and Gang Hua.
\newblock Stereoscopic neural style transfer.
\newblock In {\em CVPR}, 2018.

\bibitem{chen2018deeplab}
Liang-Chieh Chen, George Papandreou, Iasonas Kokkinos, Kevin Murphy, and Alan~L
  Yuille.
\newblock Deeplab: Semantic image segmentation with deep convolutional nets,
  atrous convolution, and fully connected crfs.
\newblock {\em TPAMI}, 2018.

\bibitem{gong2017motion}
Dong Gong, Jie Yang, Lingqiao Liu, Yanning Zhang, Ian~D Reid, Chunhua Shen,
  Anton Van Den~Hengel, and Qinfeng Shi.
\newblock From motion blur to motion flow: A deep learning solution for
  removing heterogeneous motion blur.
\newblock In {\em CVPR}, 2017.

\bibitem{gong2018neural}
Xinyu Gong, Haozhi Huang, Lin Ma, Fumin Shen, Wei Liu, and Tong Zhang.
\newblock Neural stereoscopic image style transfer.
\newblock In {\em ECCV}, 2018.

\bibitem{hirsch2011fast}
Michael Hirsch, Christian~J Schuler, Stefan Harmeling, and Bernhard Scholkopf.
\newblock Fast removal of non-uniform camera shake.
\newblock In {\em ICCV}, 2011.

\bibitem{hu2014joint}
Zhe Hu, Li Xu, and Ming-Hsuan Yang.
\newblock Joint depth estimation and camera shake removal from single blurry
  image.
\newblock In {\em CVPR}, 2014.

\bibitem{hyun2017online}
Tae Hyun~Kim, Kyoung Mu~Lee, Bernhard Scholkopf, and Michael Hirsch.
\newblock Online video deblurring via dynamic temporal blending network.
\newblock In {\em CVPR}, 2017.

\bibitem{ilg2018occlusions}
Eddy Ilg, Tonmoy Saikia, Margret Keuper, and Thomas Brox.
\newblock Occlusions, motion and depth boundaries with a generic network for
  disparity, optical flow or scene flow estimation.
\newblock In {\em ECCV}, 2018.

\bibitem{jeon2018enhancing}
Daniel~S Jeon, Seung-Hwan Baek, Inchang Choi, and Min~H Kim.
\newblock Enhancing the spatial resolution of stereo images using a parallax
  prior.
\newblock In {\em CVPR}, 2018.

\bibitem{johnson2016perceptual}
Justin Johnson, Alexandre Alahi, and Li Fei-Fei.
\newblock Perceptual losses for real-time style transfer and super-resolution.
\newblock In {\em ECCV}, 2016.

\bibitem{kim2018spatio}
Tae~Hyun Kim, Mehdi~SM Sajjadi, Michael Hirsch, and Bernhard Sch{\"o}lkopf.
\newblock Spatio-temporal transformer network for video restoration.
\newblock In {\em ECCV}, 2018.

\bibitem{kingma2015adam}
Diederik~P Kingma and Jimmy Ba.
\newblock Adam: A method for stochastic optimization.
\newblock In {\em ICLR}, 2015.

\bibitem{kupyn2018deblurgan}
Orest Kupyn, Volodymyr Budzan, Mykola Mykhailych, Dmytro Mishkin, and Jiri
  Matas.
\newblock Deblurgan: Blind motion deblurring using conditional adversarial
  networks.
\newblock In {\em CVPR}, 2018.

\bibitem{lee2017joint}
Dongwoo Lee, Haesol Park, In~Kyu Park, and Kyoung~Mu Lee.
\newblock Joint blind motion deblurring and depth estimation of light field.
\newblock In {\em ECCV}, 2018.

\bibitem{li2018depth}
Bing Li, Chia-Wen Lin, Boxin Shi, Tiejun Huang, Wen Gao, and C-C~Jay Kuo.
\newblock Depth-aware stereo video retargeting.
\newblock In {\em CVPR}, 2018.

\bibitem{li2019blind}
Lerenhan Li, Jinshan Pan, Wei-Sheng Lai, Changxin Gao, Nong Sang, and
  Ming-Hsuan Yang.
\newblock Blind image deblurring via deep discriminative priors.
\newblock {\em IJCV}, 2019.

\bibitem{li2010generating}
Yunpeng Li, Sing~Bing Kang, Neel Joshi, Steve~M Seitz, and Daniel~P
  Huttenlocher.
\newblock Generating sharp panoramas from motion-blurred videos.
\newblock In {\em CVPR}, 2010.

\bibitem{liu2016learning}
Sifei Liu, Jinshan Pan, and Ming-Hsuan Yang.
\newblock Learning recursive filters for low-level vision via a hybrid neural
  network.
\newblock In {\em ECCV}, 2016.

\bibitem{mayer2016large}
Nikolaus Mayer, Eddy Ilg, Philip Hausser, Philipp Fischer, Daniel Cremers,
  Alexey Dosovitskiy, and Thomas Brox.
\newblock A large dataset to train convolutional networks for disparity,
  optical flow, and scene flow estimation.
\newblock In {\em CVPR}, 2016.

\bibitem{moritz2015kitti}
Moritz Menze and Andreas Geige.
\newblock Object scene flow for autonomous vehiclesk.
\newblock In {\em CVPR}, 2015.

\bibitem{nah2017deep}
Seungjun Nah, Tae~Hyun Kim, and Kyoung~Mu Lee.
\newblock Deep multi-scale convolutional neural network for dynamic scene
  deblurring.
\newblock In {\em CVPR}, 2017.

\bibitem{niklaus2017iccv}
Simon Niklaus, Long Mai, and Feng Liu.
\newblock Video frame interpolation via adaptive separable convolution.
\newblock In {\em ICCV}, 2017.

\bibitem{noroozi2017motion}
Mehdi Noroozi, Paramanand Chandramouli, and Paolo Favaro.
\newblock Motion deblurring in the wild.
\newblock In {\em GCPR}, 2017.

\bibitem{pan2016blind}
Jinshan Pan, Deqing Sun, Hanspeter Pfister, and Ming-Hsuan Yang.
\newblock Blind image deblurring using dark channel prior.
\newblock In {\em CVPR}, 2016.

\bibitem{pan2017simultaneous}
Liyuan Pan, Yuchao Dai, Miaomiao Liu, and Fatih Porikli.
\newblock Simultaneous stereo video deblurring and scene flow estimation.
\newblock In {\em CVPR}, 2017.

\bibitem{pang2018zoom}
Jiahao Pang, Wenxiu Sun, Chengxi Yang, Jimmy Ren, Ruichao Xiao, Jin Zeng, and
  Liang Lin.
\newblock Zoom and learn: Generalizing deep stereo matching to novel domains.
\newblock In {\em CVPR}, 2018.

\bibitem{paramanand2013non}
Chandramouli Paramanand and Ambasamudram~N Rajagopalan.
\newblock Non-uniform motion deblurring for bilayer scenes.
\newblock In {\em CVPR}, 2013.

\bibitem{park2017joint}
Haesol Park and Kyoung~Mu Lee.
\newblock Joint estimation of camera pose, depth, deblurring, and
  super-resolution from a blurred image sequence.
\newblock In {\em ICCV}, 2017.

\bibitem{pytorch}
Adam Paszke, Sam Gross, Soumith Chintala, Gregory Chanan, Edward Yang, Zachary
  DeVito, Zeming Lin, Alban Desmaison, Luca Antiga, and Adam Lerer.
\newblock Automatic differentiation in pytorch.
\newblock In {\em NIPS Workshops}, 2017.

\bibitem{sellent2016stereo}
Anita Sellent, Carsten Rother, and Stefan Roth.
\newblock Stereo video deblurring.
\newblock In {\em ECCV}, 2016.

\bibitem{simonyan2015very}
Karen Simonyan and Andrew Zisserman.
\newblock Very deep convolutional networks for large-scale image recognition.
\newblock In {\em ICLR}, 2015.

\bibitem{su2017deep}
Shuochen Su, Mauricio Delbracio, Jue Wang, Guillermo Sapiro, Wolfgang Heidrich,
  and Oliver Wang.
\newblock Deep video deblurring for hand-held cameras.
\newblock In {\em CVPR}, 2017.

\bibitem{sun2015learning}
Jian Sun, Wenfei Cao, Zongben Xu, and Jean Ponce.
\newblock Learning a convolutional neural network for non-uniform motion blur
  removal.
\newblock In {\em CVPR}, 2015.

\bibitem{sundaram2010dense}
Narayanan Sundaram, Thomas Brox, and Kurt Keutzer.
\newblock Dense point trajectories by gpu-accelerated large displacement
  optical flow.
\newblock In {\em ECCV}, 2010.

\bibitem{tao2018scale}
Xin Tao, Hongyun Gao, Xiaoyong Shen, Jue Wang, and Jiaya Jia.
\newblock Scale-recurrent network for deep image deblurring.
\newblock In {\em CVPR}, 2018.

\bibitem{whyte2012non}
Oliver Whyte, Josef Sivic, Andrew Zisserman, and Jean Ponce.
\newblock Non-uniform deblurring for shaken images.
\newblock {\em IJCV}, 2012.

\bibitem{xu2012depth}
Li Xu and Jiaya Jia.
\newblock Depth-aware motion deblurring.
\newblock In {\em ICCP}. IEEE, 2012.

\bibitem{xu2013unnatural}
Li Xu, Shicheng Zheng, and Jiaya Jia.
\newblock Unnatural l0 sparse representation for natural image deblurring.
\newblock In {\em CVPR}, 2013.

\bibitem{lecun2015stereo}
Jure Zbontar and Yann LeCun.
\newblock Computing the stereo matching cost with a convolutional neural
  network.
\newblock In {\em CVPR}, 2015.

\bibitem{zhang2017learning}
Jiawei Zhang, Jinshan Pan, Wei-Sheng Lai, Rynson~WH Lau, and Ming-Hsuan Yang.
\newblock Learning fully convolutional networks for iterative non-blind
  deconvolution.
\newblock In {\em CVPR}, 2017.

\bibitem{zhang2018dynamic}
Jiawei Zhang, Jinshan Pan, Jimmy Ren, Yibing Song, Linchao Bao, Rynson~WH Lau,
  and Ming-Hsuan Yang.
\newblock Dynamic scene deblurring using spatially variant recurrent neural
  networks.
\newblock In {\em CVPR}, 2018.

\bibitem{zoran2011learning}
Daniel Zoran and Yair Weiss.
\newblock From learning models of natural image patches to whole image
  restoration.
\newblock In {\em ICCV}, 2011.

\end{thebibliography}
}

\end{document}